\documentclass{article} 
\usepackage{iclr2025_conference,times}

\usepackage{amsmath,amsfonts,bm}

\def\eqref#1{equation~\ref{#1}}

\def\1{\bm{1}}

\def\vb{{\bm{b}}}

\def\vv{{\bm{v}}}

\def\vx{{\bm{x}}}

\def\mG{{\bm{G}}}
\def\mH{{\bm{H}}}
\def\mI{{\bm{I}}}

\def\mM{{\bm{M}}}

\def\mW{{\bm{W}}}

\DeclareMathAlphabet{\mathsfit}{\encodingdefault}{\sfdefault}{m}{sl}
\SetMathAlphabet{\mathsfit}{bold}{\encodingdefault}{\sfdefault}{bx}{n}

\usepackage[utf8]{inputenc} 
\usepackage[T1]{fontenc}    
\usepackage{hyperref}       
\usepackage{url}            
\usepackage{booktabs}       
\usepackage{amsfonts}       
\usepackage{nicefrac}       
\usepackage{microtype}      
\usepackage{xcolor}         
\usepackage{caption}
\usepackage{subcaption}
\usepackage{graphicx}
\usepackage{amsmath}
\usepackage{arydshln} 
\usepackage{bm} 
\usepackage{lipsum}
\usepackage{amsthm}
\usepackage{amssymb}
\usepackage{mathtools}
\usepackage{cancel}

\title{Can Stability be Detrimental? \\Better Generalization through Gradient \\Descent Instabilities}

\author{Lawrence Wang, Stephen J. Roberts \\
Department of Engineering\\
University of Oxford\\
Oxford, UK \\
\texttt{\{lawrence.wang@exeter.ox.ac.uk, sjrob@robots.ox.ac.uk}
}

\iclrfinalcopy

\begin{document}

\newcommand{\fcite}{\textcolor{red}{cite}}
\newcommand{\lammax}{$\lambda_\mathrm{max}$}
\newcommand{\todo}{\textcolor{red}{todo}}

\newcommand*{\crosssymbol}{%
    \text{%
      \raisebox{1ex}{%
        \makebox[0pt][l]{%
          \rule[-.2pt]{.75ex}{.4pt}%
        }%
        \makebox[.75ex]{%
          \rule[-1ex]{.4pt}{1.5ex}%
        }%
      }%
    }%
}   

\makeatletter
\newcommand\footnoteref[1]{\protected@xdef\@thefnmark{\ref{#1}}\@footnotemark}
\makeatother

\newtheorem*{definition}{Definition}
\newtheorem*{remark}{Remark}
\newtheorem*{lemma}{Lemma}

\maketitle

\begin{abstract}


Traditional analyses of gradient descent optimization show that, when the largest eigenvalue of the loss Hessian - often referred to as the sharpness - is below a critical learning-rate threshold, then training is `stable' and training loss decreases monotonically. Recent studies, however, have suggested that the majority of modern deep neural networks achieve good performance despite operating outside this stable regime. In this work, we demonstrate that such instabilities, induced by large learning rates, move model parameters toward flatter regions of the loss landscape. Our crucial insight lies in noting that, during these instabilities, the orientation of the Hessian eigenvectors rotate. This, we conjecture, allows the model to explore regions of the loss landscape that display more desirable geometrical properties for generalization, such as flatness. These rotations are a consequence of network depth, and we prove that for any network with depth $>1$, unstable growth in parameters causes rotations in the principal components of the Hessian, which promotes exploration of the parameter space away from unstable directions. Our empirical studies reveal an implicit regularization effect in gradient descent with large learning rates operating beyond the stability threshold. We find these lead to improved generalization performance on modern benchmark datasets. 
\end{abstract}
\section{Introduction}\label{sec:intro}
Deep neural networks are widely successful across a number of tasks, but their generalization performance is dependent on careful choices of hyperparameters which govern the learning process. Gradient descent (including stochastic gradient descent and ADAM \citep{kingma2017adammethodstochasticoptimization}) is arguably the most widely-used learning algorithm due to its simplicity and versatility. For such methods, the \emph{descent lemma} upper-bounds the choice of learning rate by the local curvature (or sharpness) to guarantee stable optimization trajectories and provable decreases for convex training losses.

Recently, the `unstable' learning-rate regime has been a focal point for research. \cite{cohen2022gradientdescentneuralnetworks} have demonstrated that in practice, learning rates can go above the stability threshold as determined by the \emph{descent lemma}. Surprisingly, this appears not to destabilize training trajectories as expected, but instead model sharpness continues near the stability limit while training loss improves. Large learning rates have also been known to improve generalization performance. Our work builds on these findings, showing that training instabilities, caused by large learning rates, drive model parameters toward flatter regions of the loss landscape to improve generalization. 



The main contribution of our work, detailed in Section \ref{sec:rot}, proposes that these instabilities are resolved through rotations in the eigenvectors of the loss Hessian. We demonstrate that, for deep neural networks, unstable training causes these eigenvectors to rotate away from their previous orientations, whereas in the stable regime, the orientations are reinforced. These rotations allow the model to explore regions of parameter space with an in-built bias for flatness. We term the accumulation of this implicit regularization effect \emph{progressive flattening}, which we validate empirically. 


Throughout our experiments, learning rates allow direct control over the magnitude of these regularization effects. Our empirical study in Section \ref{sec:generalization} explores the relationship between learning rates and generalization, revealing a clear phase transition where generalization benefits only emerge for learning rates beyond the stability threshold, solidifying the role of instabilities in regularization. Additionally, we show that starting with large learning rates have long-term benefits in generalization performance, even when learning rates are reduced later in training. However, our findings also challenge the reliability of sharpness as a metric for generalization, as the degree of eigenvector rotation can, in some cases, be a more effective predictor. The code for reproducing the experiments and results in this paper will be made available on GitHub upon publication.

\section{Background}\label{sec:bg}

\subsection{Gradient Descent and the Descent Lemma}\label{sec:bg:gd}
We denote the loss function by $L(\theta)$ with parameters $\theta$ that are updated with gradient descent using learning rate $\eta$, i.e. $\theta_{t+1} = \theta_t - \eta \nabla L(\theta_t)$. The so-called \textit{sharpness} $S$ of the loss landscape is typically estimated with the maximum curvature of the loss Hessian $H (\theta)\coloneq \nabla^2 L(\theta)$, through the maximum eigenvalue $S(\theta) = \lambda_\mathrm{max}(H(\theta))$. The \textit{descent lemma} can be stated as: 

\begin{lemma}
\text{For a convex, $l$-smooth function $L(\theta)$, } $L(\theta_{t+1}) \leq L(\theta_t) - \eta(1-\frac{\eta l}{2})|{\nabla L(\theta_t)}|^2_2$
\end{lemma}
The proof uses the co-coercive property of $l$-smooth functions using the results of \cite{Baillon1977QuelquesPD}. The decrease in loss is scaled by the quadratic $\eta(1- \frac{\eta l}{2})$, which leads to the optimal learning rate $\eta=1/l$. However, any choice of $0<\eta<2/l$ guarantees a decrease in the loss function, allowing convergence to the minima in the \emph{stable} regime of training, leading to a popular rule for $\eta$ selection. On the other hand, choosing $\eta>2/l$ results in so-called \emph{instabilities}, causing $L$ and $\theta$ to grow without bound. Additionally, when $\eta>1/l$, parameters \emph{oscillate} about the minima. 

Although these bounds are derived from $l$, in practice $\eta$ is chosen without knowing $l$, leading to an empirical stability threshold of $S(\theta)$ given by $S(\theta) \leq 2/\eta$, beyond which training is thought to destabilize. However, \cite{cohen2022gradientdescentneuralnetworks} showed that in practice, training is not destabilized as expected. They identified two important phenomena for gradient descent: 

\textbf{Progressive Sharpening}: So long as training is stable ($S(\theta) \leq 2/\eta$), gradient descent has an overwhelming tendency to continually increase sharpness. 

\textbf{Edge of Stability}: Once sharpness reaches the stability limit, it sits at, or just above, the stability threshold. Additionally, although the descent lemma does not guarantee a decrease of training loss, the training loss nonetheless continues to decrease, albeit non-monotonically.

In practice, loss functions extend beyond the quadratic and training can go beyond the stability limit in phases of \emph{instability}, which manifests as spikes in $\theta$, $L(\theta)$, and $S(\theta)$. In addition to the extensive analysis of the self-stabilization property of gradient descent, \cite{damian2023selfstabilizationimplicitbiasgradient} introduced the progressive sharpening factor $\alpha=-\nabla L(\theta) \cdot \nabla S(\theta).$, where a positive $\alpha$ indicates that gradient updates promote sharpening. These authors suggested that it is sensible to assume, $\alpha > 0$, in a constrained trajectory, including during instabilities. Our exploratory experiments (see Appendix \ref{app:drivers}) suggest that this effect is not clear in the real-world scenarios that we studied, and may play a limited role. For the purposes of our study, we posit that the primary driver of instability is unstable growth in parameters, driven by $S(\theta)>2/\eta$ derived from the quadratic approximation.

\subsection{Sharpness and Generalization}\label{sec:bg:sharp}

The notion that solutions with low sharpness, or flat minima, promote generalization performance is widely accepted. \cite{SchmidhuberFlat1997} argued, using minimum description length, that flatter solutions are less complex, thereby improving generalization through appeal to \emph{Occam's Razor}. More recently, \cite{keskar2017largebatchtrainingdeeplearning} and \cite{jastrzębski2018factorsinfluencingminimasgd} observed that deep neural networks trained with small learning rates tend to generalize poorly because the minima they converged to were narrow. The width of minima is measured through sharpness, $S(\theta)$, often defined as the largest eigenvalue of the loss Hessian. However, \cite{dinh2017sharpminimageneralizedeep} demonstrated that some output-preserving transformations can lead to arbitrary values of $S(\theta)$, revealing a lack of scale-invariance. While this observation weakens the absolute causality from sharpness to generalization, \cite{jiang2019fantasticgeneralizationmeasures} found that sharpness may still serve as a useful indicator of generalization performance. 

Modern methods for efficiently computing eigenvalue-vector pairs of the Hessian utilize \cite{Pearlmutter1994FastEM}'s trick, which allows evaluation of the Hessian-vector product without explicitly forming the Hessian. In our work, we present an implementation in \emph{Jax} \citep{jax2018github} that also employs the Modified Parlett-Kahan re-orthogonalization \citep{Abaffy20152TP} to enhance numerical stability. Additionally, our study focuses on the similarity between eigenvectors, which can naturally be compared with cosine-similarity in one-dimension. For comparisons between subspaces formed by multiple eigenvectors, we use the cosine-Grassmanian distance \citep{ye2016schubertvarietiesdistancessubspaces}. The code to reproduce the results in this paper will be made available on GitHub upon publication.

\section{Regularization through Instabilities} \label{sec:rot}

We study the dynamics of gradient descent during instabilities - a phase of learning driven by unstable parameter growth and characterized by spikes in the parameter $\theta$, the loss function $L(\theta)$ and sharpness $S(\theta)$. In this section, we introduce a toy problem to demonstrate that UPG causes rotations in the sharp eigenvectors of the loss Hessian. 

Adopting a reductionist perspective, our overall approach attempts to explain the complex behavior of the loss function by analyzing an individual term, represented by a Diagonal Linear Network (DLN). Our approach is motivated by the Sum of Multiplications (SoM) decomposition on the outputs of a multilayer perceptron (MLP), which we motivate briefly.

In general, let the outputs of an MLP be:
\begin{align}
    f(\vx) &= \mW_d \sigma_{d-1} (\mW_{d-1} \sigma_{d-2} ( ... \sigma_1(\mW_1 \vx + \vb_1) ... ) + \vb_{d-1} ) +  \vb_d \nonumber
\end{align}
where the $p$-dimensional inputs to the neural network $\vx \in \mathbb{R}^{h_0} = \mathbb{R}^{p}$. Each hidden-layer of index $i, i \in 1, 2, ... d$ has input and output dimensions $\mathbb{R}^{h_{i-1}} \rightarrow \mathbb{R}^{h_{i}}$, weights $\mW_i\in \mathbb{R}^{h_i \times h_{i-1}}$, and bias terms $\vb_i \in \mathbb{R}^{h_i}$. These layers are followed by the activation functions $\sigma_{i}$. A single-output network ($h_d=1$) with identity activations ($\sigma=\mI$) can be reduced to a summation of multiplications form, where the network output can be seen as a \emph{summation over multiplicative} terms (SoM). (For details, see Appendix \ref{app:dln-reality}). 

To model the dynamics of one such term, we utilize a DLN, a simple neural network with non-trivial dynamics, as studied in \cite{pesme2021implicitbiassgddiagonal}. The DLN's multiplicative structure reflects a key feature of depth, which weights are multiplied across vertically-stacked neural layers whenever depth $d>1$. We focus on a detailed study of a simple $2$-parameter $1$-DLN model for an MLP, where computations are exact to illustrate key insights. With this study, we identify phases of $\eta$ that match up exceedingly well with existing stability bounds, with clear rotational biases for each of these phases. We then provide empirical validation for our rotational claims, before combining our insights with an empirical study of instability resolution, where we form a conjecture on the mechanism through which training returns to stability. Finally, we offer observations of \emph{progressive flattening}, where repeated instabilities lead to generally flatter minima in the loss landscape. 

Our theoretical studies in this section use simplifying assumptions, i.e. $n=2$, $\sigma=\mI$, and $m=1$. We provide derivations for general $n$ in Appendix \ref{app:dln_full}; we extend our results to the ReLU activation in Appendix \ref{app:dln-relu}; and finally, Appendix \ref{app:dln-loss} considers the extension of our claims to a general $n$-parameter $m$-DLN SoM formulation to polynomial loss functions, not just simply the outputs of the MLP as we have motivated here.

\subsection{Rotations from Parameter Growth - a DLN model} \label{sec:rot:dln2}

The outputs of a DLN, $f(\theta_1, ... \theta_n) = \Theta = \prod ^n_i {\theta_i}$ is a multiplication of individual weights. For any neural network with depth $d>1$, the SoM decomposition of the outputs involve individual terms that reflect the product of weights across layers, whose individual path-structure is captured by the DLN formulation. We focus on one such term, $\Theta = \prod ^n_i {\theta_i}$, and we demonstrate that parameter growth induces rotations in the sharpest Hessian eigenvectors in this simple multiplicative formulation. By examining the two-parameter case, where exact derivations are possible, we obtain key insights into these rotations. The extension to a general $n$-parameter DLN is explored in detail in Appendix \ref{app:dln_full}. 

Let the loss $L(\Theta)$ be described by $z(\Theta)$, a nonnegative convex polynomial of degree $q$ on the network outputs $\Theta$ with a unique minimum at $\Theta=\theta_1\theta_2=0$, which limits $z(\Theta)$ to even degree polynomials. Additionally, assume that the parameters are not at the minimum, i.e. $\Theta \neq 0$. Writing $z':=\frac{\partial z}{\partial \Theta}$, we get the loss Hessian $\mH(\Theta)$:
\begin{align}
\mH&= \begin{bmatrix}
    \frac{\partial^2 L}{\partial \theta_1^2} & \frac{\partial^2 L}{\partial \theta_1 \theta_2} \\
    \frac{\partial^2 L}{\partial \theta_1 \theta_2} & \frac{\partial^2 L}{\partial \theta_2^2} \\
    \end{bmatrix} = \begin{bmatrix}
    z'' \theta_2^2 & z'' \theta_1 \theta_2 +z' \\
    z'' \theta_2 \theta_1 +z' & z'' \theta_1^2 \\
    \end{bmatrix} \nonumber
\intertext{where the dependence of $H$, $L$, and $z$ on the inputs $\Theta$ has been dropped for notational clarity. Additionally, denote the eigenvector-value pairs $(\lambda_i, \vv_i), i\in \{ 1, 2 \}$ in the basis formed by $\theta$s: $\vv_i := \begin{bmatrix}u_{i, 1},u_{i, 2}\end{bmatrix}^T$. We solve exactly for $n=2$ to get $R_2$, a ratio of coordinates for $n=2$ (see Appendix \ref{app:dln2-beta}):}
    R_2 &= \left|\frac{u_{1, 1}}{u_{1, 2}}\right| = \beta + \sqrt{\beta^2+1}, \text{ where } \beta=\left|\frac{z''(\theta_2^2-\theta_1^2)}{2(z'+z''\Theta)}\right|
\intertext{We note that $R_2$ is a monotonic function of $\beta$. Practically, $R_2$ reflects a degree of alignment between the sharpest eigenvector and the sharpest parameter, which we will use to characterize the orientation of the sharpest eigenvector following gradient updates.}
\intertext{Consider an optimization trajectory initialized at $\begin{bmatrix}\theta_{1},\theta_{2}\end{bmatrix}^T$ using a fixed learning rate $\eta$. Without loss of generality, assume that $\theta_2^2 > \theta_1^2$, which implies that $L(\Theta)$ is more sensitive to changes in $\theta_1$ than $\theta_2$, or that $\theta_1$ is the sharper parameter. Importantly, we note that $R_2$ is sign-invariant (see Appendix \ref{app:dln2-beta}) and is invariant to exchanges in the magnitude of $\theta$s. The gradient updates of $\theta$s are:}
    \Delta \theta_1 &= \eta \frac{dz}{d\theta_1} = \eta z' \theta_2; \text{ similarly } \Delta \theta_{2} =  \eta z' \theta_1 \nonumber
\intertext{Despite originating from the same loss function $z$, the updates to these parameters vary in scale owing to the multiplicative nature of $\Theta$. This leads to $\Delta \theta_1>\Delta\theta_2$, leading to instabilities when $\eta$ is too large. Let $\gamma_x$ denote the ratio of changes in variable $x$. Given these updates, $\gamma_\beta$ is:}
    \gamma_\beta := \frac{\beta + \Delta \beta}{\beta} &\approx \frac{\gamma_{\left|\theta_2^2-\theta_1^2\right|}}{\gamma_\Theta} = \left|\frac{1-\eta^2z'^2}{1-s\eta z'+\eta^2z'^2}\right| \text{ where } s=\left (\frac{\theta_1}{\theta_2}+\frac{\theta_2}{\theta_1}\right ) > 2
\end{align}

The regimes of $\gamma_\beta$ can be studied through the behavior of the denominator, $\gamma_\Theta$. The function asymptotes when $\gamma_\Theta \rightarrow 0$ at $\eta z' \in \{\theta_1/\theta_2, \theta_2/\theta_1 \}$. The (left-side) asymptote corresponding to the smaller value of $\eta$ occurs when $\theta_1=0$ after one update. For quadratic $z$, the learning rate $\eta = \theta_2/(\theta_1 z') = 1/\theta_1^2 = 1/(z''\theta_2^2)$, is exactly half of the stability limit when considering the local quadratic approximation with respect to $\theta_1$, which reveals a sensible but \emph{non-trivial} connection. The existence of the first asymptote within the stable regime of learning rates is therefore guaranteed through this connection. Later, the function reaches $\gamma_\beta=1$ at:
\begin{align*}
    \eta_{\gamma_\beta=1} z' &= \frac{2\theta_1\theta_2}{\theta_1^2 + \theta_2^2} = \frac{2\theta_1}{\theta_2}\left (1-\frac{\theta_1^2}{\theta_1^2+\theta_2^2} \right ) = \eta_\mathrm{eos}(1-\xi) 
\end{align*}
where we defined $\xi = \theta_1^2/(\theta_1^2+\theta_2^2)$, which is small as $\theta_2^2>\theta_1^2$. 

\subsection{Rotational Polarity of Eigenvectors}\label{sec:rpe}

\begin{figure}[t]
\includegraphics[width=1\textwidth]{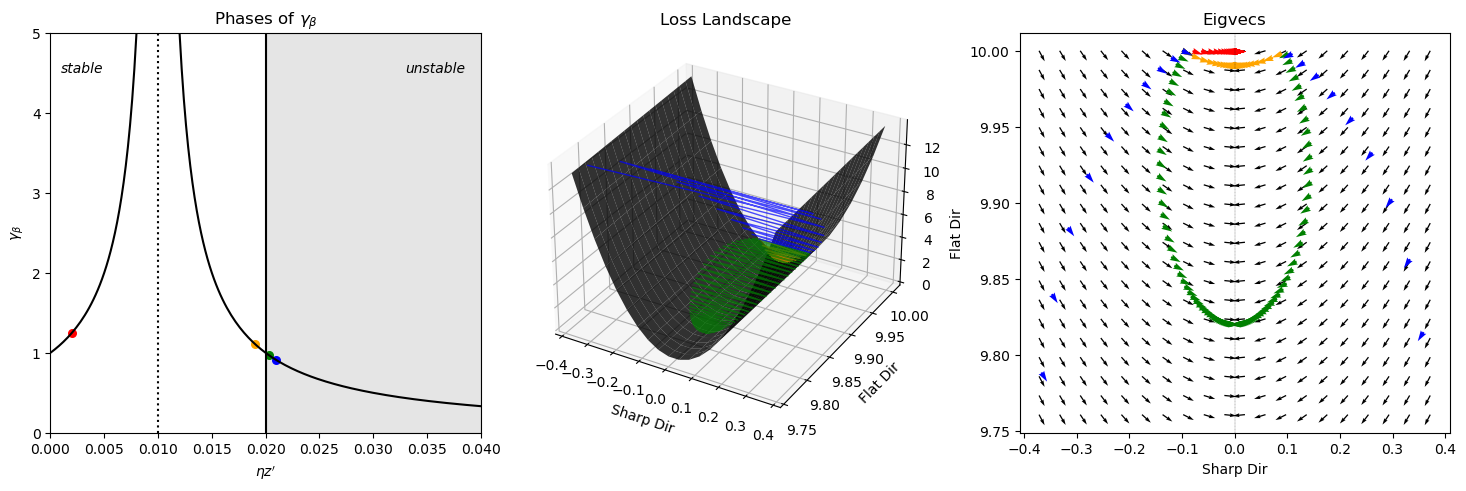}
\caption{\textbf{Optimization trajectories in a $2$-parameter DLN display rotations. } We visualize $4$ our trajectories, initialized at $(-0.1, 10)$ with $z = \Theta^2$ and choosing $\eta \in \{0.001, 0.0095, 0.011, 0.013\}$, where the stability limit (at initialization) is $\eta_\mathrm{eos} = 0.01$. \textbf{Left:} the regimes of $\gamma_\beta$. \textbf{Middle:} the loss landscape. \textbf{Right:} map of eigenvector orientations (y-axis magnitudes amplified for clarity). }
\label{fig:cartoon}
\end{figure}

Following the previous analysis, we characterize the regimes of $\gamma_\beta$ as a function of $\eta$: 
\begin{enumerate}
    \item From $ 0 \leq \eta \leq \frac{2\theta_1}{\theta_2 z'}(1-\xi)$ is the \textbf{stable} regime of training\footnote{\label{note: 2lam} When $z$ is a quadratic function, $\frac{\theta_1}{\theta_2 z'}=1/\lambda$, so the above thresholds are equal to the $1/\lambda$ and $2/\lambda$ bounds.}. In these regimes, $\gamma_\beta>1$, which leads to increases in $R_2$, signifying an increased alignment of the sharpest eigenvector $v_1$ to $\theta_1$, the sharper parameter. As $\eta z' \rightarrow \frac{\theta_1}{\theta_2}$, $\gamma_\Theta \rightarrow \infty$. 
    \item As $\eta > \frac{2\theta_1}{\theta_2 z'}(1-\xi)$, we enter the \textbf{unstable} regime. In this regime, $\gamma_\beta<1$, leading to decreases in $R_2$, implying that the alignment of the sharpest eigenvector $v_1$ rotates away from $\theta_1$, the sharper parameter. 
\end{enumerate}

The existence of the unstable regime along $\theta_1$ requires that the assumption on relative magnitudes is upheld, i.e. $\eta_\mathrm{eos} < \eta_{\gamma_\beta=0} = 1/z' \implies \theta_2^2>2\theta_1^2$, which is a mild constraint on the relative magnitudes of parameters, especially when considering the ill-conditioning of deep neural networks in practice  \citep{papyan2019measurementsthreelevelhierarchicalstructure}. In this practical regime, $\xi \approx 0$, so that the threshold for regime change is exactly equal to the EoS limit for quadratic $z$. In Figure \ref{fig:cartoon}, we observe different behaviors in the orientation of $\vv_1$ for different choices of $\eta$, using quadratic $z$. The $\eta=[0.001, 0.0095]$ (red and orange respectively) minima represent the attractors accessible to different choices of $\eta$s as $\theta_1 \rightarrow 0$, with the same orientations of $\vv_1$. For unstable $\eta=[0.011, 0.013]$, both trajectories show $\vv_1$ initially moving away from $\theta_1$, but return to stability later on. These differences in eigenvector rotations owing to the stability of $\eta$ reveal the  \textbf{rotational polarity of eigenvectors (RPE)} in gradient descent - eigenvectors rotate away from parameter orientations that are \emph{too sharp} while parameter orientations that are \emph{flat enough} under the stability limit may be attractors to the trajectory. This is the main claim of our theoretical analyses, and is crucially important to the our conjectures on instability resolution. 

\begin{figure}[t] 
\includegraphics[width=\linewidth]{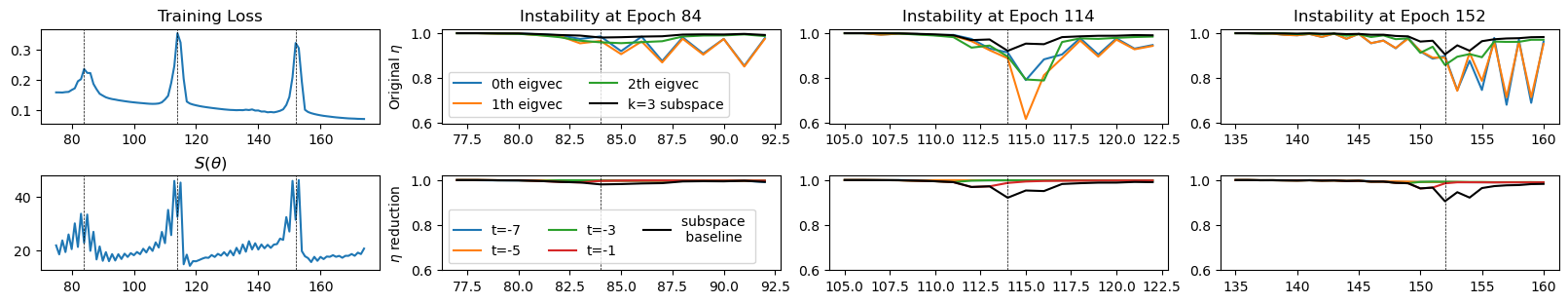}
\caption{\textbf{During instabilities, the sharpest eigenvectors of the Hessian rotate away smoothly and monotonically (Top), while stable training reverts these rotations (Bottom).} We track the similarity of the sharpest Hessian eigenvectors across epochs through three instabilities. \textbf{Left:} $L(\theta)$ and $S(\theta)$. \textbf{Top:} similarities of the $k$-th eigenvectors (colored) and of subspaces formed by the top $3$ eigenvectors (black) during instabilities. \textbf{Bottom:} similarity of subspaces formed by the top $3$ eigenvectors to the baseline (black) across various timings (colored) when $\eta$ reduction begins. }
\label{fig:instab_rot}
\end{figure}

Common to both regimes of $\eta$ we observe the monotonicity of rotations, where stable $\eta$s directly rotate toward $\theta_1$, while unstable $\eta$s rotate away, at least initially. We train an MLP on fMNIST \citep{xiao2017fashionmnistnovelimagedataset}, and we track the rotation of the sharpest Hessian eigenvectors in Figure \ref{fig:instab_rot}. In this figure, we plot the degree of rotation among the sharpest Hessian eigenvectors at three snapshots of instability, where we compared eigenvector similarities to baselines defined at the beginning of each snapshot. Focusing on the top panel, we observe gradual and monotonic rotations as $L(\theta)$ approaches its peak values as predicted by theory. Notably, these rotations do not involve sudden changes in any single eigenvector but reflect a general decrease in similarity across all eigenvectors, conforming to our expectations. Interestingly, even after the instability is resolved, the similarity among individual eigenvectors fall while the overall subspace similarity remains high. Since the subspace of top eigenvectors is heavily constrained by the problem \citep{papyan2019measurementsthreelevelhierarchicalstructure}, the re-orientation of individual sharp eigenvectors suggests that novel combinations of these eigenvectors can be beneficial toward flatness. In the bottom pane, we intervene to enforce stability by setting $\eta_\mathrm{low} = 0.2\eta$s at different time-steps before the instability is resolved. Since these models have undergone some instability, the eigenvectors have rotated away from their stable orientations (akin to the $\theta_1$ direction in the $2$-parameter DLN), but these rotations are immediately reversed once $\eta$-reduction takes place. These observations support the theoretical claims of RPE and monotonicity. 

For both the theoretical DLN model and the real-world MLP, we observe an eventual resolution of instability with unstable $\eta$s. Resolution in the DLN case is driven by an improvement (reduction) in the parameter $\Theta$ through $\theta_2$, thereby reducing the overall sharpness of the landscape. This phenomenon is also predicted by the general $n$-parameter model (see Eqn. \ref{eqn:R-general}). Additionally, the predicted trajectory of resolution is smooth, but the empirical trajectory of resolution in Figure \ref{fig:instab_rot} and others are sharp. We find that this mismatch of prediction to observation may reflect a potential limitation of the DLN-SOP model - which implies characteristics of global optimization (interpreted as a reduction in $\Theta$), but is not sufficiently supported by the empirical evidence. 

As a result, the main claims of our theoretical analyses on the DLN model are concerned with RPE and initial monotonicity. In the following section, we use RPE with additional empirical studies to form a conjecture on the mechanism with which gradient descent instabilities are resolved.

\subsection{Resolution of Instabilities and Progressive Flattening} \label{sec:rot:emp}

While the rotation of eigenvectors follows predictions of the theoretical model, these predictions alone do not wholly explain the resolution of instabilities. In this section, we explore additional empirical studies, through visualization of the loss $L(\theta)$, and sharpness $S(\theta)$, landscapes through a complete cycle of learning, defined as from stability to instability and back to stability, using epochs $119$ to $158$ in the same MLP trajectory as Section \ref{sec:rpe}. While we find the evidence compelling, we note that the mechanism of eigenvector rotations does not imply a resolution of instability - a clear counter-example is to choose extremely large $\eta$s that directly lead the trajectory beyond recovery, which we dub the \emph{extremely-large} limit. Our findings are made with respect to the regime when $\eta$ is above the EoS limit but not \emph{extremely large} - this distinction is additionally useful for our subsequent findings with generalization benefits. 


\begin{figure}[!b] 
\includegraphics[width=\linewidth]{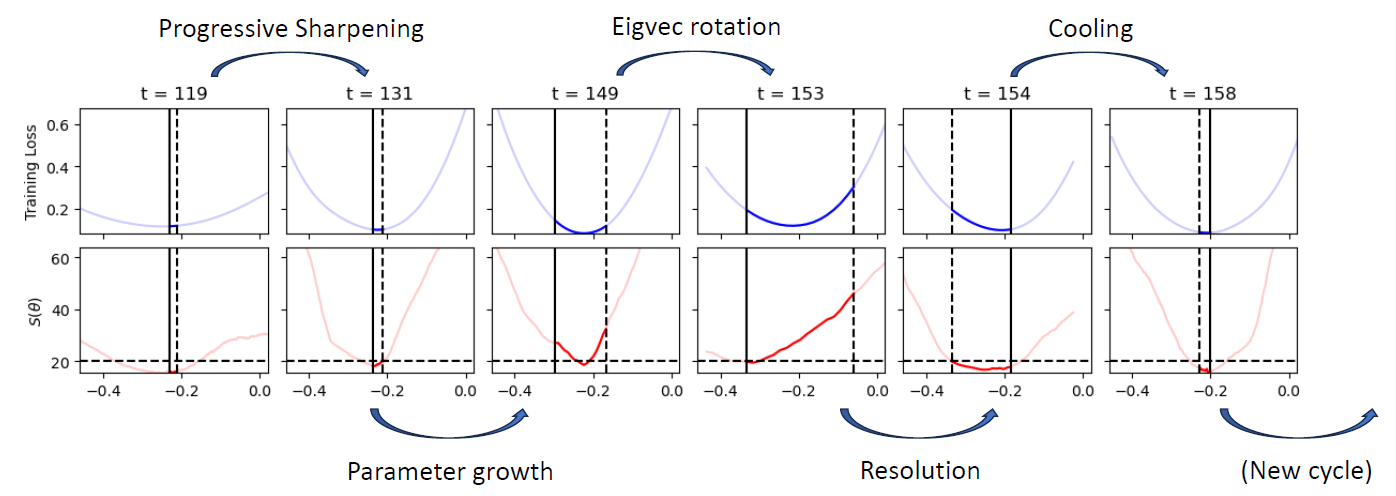}
\caption{\textbf{Parameter growth along the sharpest Hessian eigenvectors leads to exploration of the peripheries of the local minima, driving up $L(\theta)$ and $S(\theta)$ in the process. As the instability develops, the $S(\theta)$ curve undergoes large changes until a flat region is found to enable a return to stability.} We show snapshots along the instability cycle taken along the direction of the gradient. The dotted/solid vertical lines indicate the positions of previous/current parameters, respectively. }
\label{fig:movie}
\end{figure}   

As the magnitude of the parameters grow, the position of the model in the loss landscape moves beyond the region approximated by low-order Taylor Expansions at the local minima and the influence of higher-order terms are revealed. At epoch $119$, when $S(\theta)<2/\eta$, we observe a steady increase in $S(\theta)$ while $L(\theta)$ decreases. This is progressive sharpening. During this phase, the curvature of both $L(\theta)$ and $S(\theta)$ sharpens. While sharpening in $L(\theta)$ is well-documented (as an increase in $S(\theta)$), the sharpening of $S(\theta)$ indicates the increased influence of higher-order terms in the local Taylor-expansion, and that a higher-order derivative is increased (while parameters approach the minima). After epoch $131$, when $S(\theta)>2/\eta$, the curvature of both $L(\theta)$ and $S(\theta)$ remains largely constant, but parameter growth forces the model to explore wider ridges of the local minima along the unstable directions. We expect this to eventually lead to model divergence, but significant changes to the $S(\theta)$ curve are observed after epoch $149$, and fortunately a flatter curve is found by epoch $154$. At this point, the reduction in $S(\theta)$, as a result of a flatter $S(\theta)$ curve, outweighs the increased steps as a result of large parameters, resulting in optimization steps toward stability in the following epochs while the orientations of the top eigenvectors are reinforced through stability. We observe these effects as a `cooling' of $L(\theta)$ and $S(\theta)$, before progressive sharpening eventually drives the model back toward instability to repeat the cycle. The full epoch-by-epoch progression is shown in the Appendix \ref{app:movie}. 

While a simple reduction in $S(\theta)$ is sufficient for a return to stability, our empirical study revealed that instabilities may also reduce higher order derivatives (through the curvature of $S(\theta)$). As instabilities progress, the distance between current parameters $\theta$ to the `local' optimum, $\theta_*$ owing to unstable parameter growth. In a convex higher-than-quadratic approximation, when the same eigenvector is maintained, one would expect $S(\theta)$ to be even higher in by virtue of increased distance from the optimum. The effects of RPE is evaluated at these parameters which grow further, meaning that the overall curvature of $S(\theta)$ must reduce for the resolution of instabilities. Thus, contingent on non-divergence and a return to stability, repeated cycles of `surviving' instability implies that the curvature of $S(\theta)$ is likely to reduce, to capture an implicit \emph{flatness bias} of gradient descent. We term the integrated effect of this flatness bias over an entire training trajectory as \textbf{progressive flattening}. 

\begin{figure}[h]
\includegraphics[height=3cm]{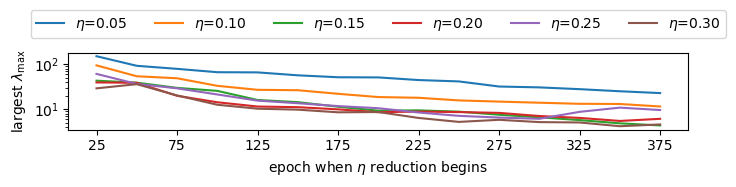}
\caption{\textbf{Progressive flattening in fMNIST.} We plot the eventual maximum $S(\theta)_\mathrm{max}$ of MLPs trained with a constant large initial learning rate $\eta_0$ before reducing to $\eta_\mathrm{small}=0.01$ at indicated epochs. The larger and longer phase with $\eta_0$, the more we observe a reduction in $S(\theta)_\mathrm{max}$. }
\label{fig:prog-flat}
\end{figure}

While we measured the curvature of curvatures for select MLPs on fMNIST, generally this is a computationally-prohibitive procedure that does not scale easily. As a proxy, we derive the extent of \emph{progressive flattening} through the reduction in progressive sharpening, whose effects on the $S(\theta)$ curve are indirect (see Appendix \ref{app:drivers}). The eventual peak in progressive sharpening, as measured in the empirical maximum sharpness $S(\theta)_\mathrm{max}$ under low-enough $\eta$s, is revealed over extended periods of stable training and indirectly reflect the curvature of $S(\theta)$ in the local landscape.  In Figure \ref{fig:prog-flat}, we track $S(\theta)_\mathrm{max}$ (for MLPs on fMNIST) as $\eta$ reductions, set to ensure stable phases of training, are applied at different stages. $S(\theta)_\mathrm{max}$ decreases as $\eta$ reductions are delayed, and larger initial $\eta_0$ used, suggesting a two-dimensional effect - both larger learning rates and prolonged training can intensify the regularization effects. Our findings highlight a strong link between training with large learning rates and the resulting flatness of local parameter space. In Section \ref{sec:generalization}, we use the connection between $S(\theta)$ and generalization to show that large learning rates contribute to improved generalization through \emph{progressive flattening}. 

\section{Effect on Generalization Performance}\label{sec:generalization}
Section \ref{sec:rot} highlighted the rotational behavior of the sharpest eigenvalues of the Hessian, which revealed an implicit bias within gradient descent instabilities for flat eigenvectors. This culminated in \emph{progressive flattening}, the accumulation of repeated flatness biases over cycles of instability, whose overall impact can be increased with larger learning rates and/or with longer durations. In this section, we present an empirical study on the benefit of large learning rates toward generalization. 

Generalization refers to the ability of neural networks to perform well on data not used in training. The generalization gap is defined as the difference between performance on training data (in-sample) and on unseen test data (out-of-sample). To standardize our comparisons, the models in this section are trained to completion, defined as achieving $>99.99\%$ accuracy on the training set. Consequently, test accuracy serves as a direct indicator of the generalization gap. 

Some of our experiments in this section are conducted on the CIFAR10 image classification dataset \citep{krizhevsky2009learning} using small VGG \citep{simonyan2015deepconvolutionalnetworkslargescale} networks. While the instability phenomena occur for many choices of error functions \citep{cohen2022gradientdescentneuralnetworks}, we found training to be more stable with cross-entropy loss compared to mean squared error loss, which motivates the use of cross-entropy loss in our experiments. These experiments are performed in a fully non-stochastic setting, with full-batch gradient descent and eschewing common data augmentation techniques such as random flips and crops \citep{AlexNet}. Additionally, since batch normalization \citep{ioffe2015batchnormalizationacceleratingdeep} benefits deep convolutional architectures, we use the non-stochastic \emph{GhostBatchNorm} \citep{hoffer2018trainlongergeneralizebetter} computed over fixed batch size $1024$, maintaining the default ordering of data in CIFAR10. To ensure the robustness of our study, we trained hundreds of models, but due to budget constraints our experiments in this section are conducted on a reduced $5$k ($10\%$) subset of the full dataset. In Section \ref{sec:gen:full-cifar}, we remove these constraints and benchmark our observations on the full $50$k dataset, and introduce non-stochastic augmentations to achieve performance near the state-of-the-art. For further details of our network architecture and experimental setup, see Appendix \ref{app:tech}. 


\subsection{Large Learning Rates}\label{sec:gen:exp1-const}

\begin{figure}[b]
    \centering
    \begin{subfigure}[t]{0.45\textwidth}
        \centering
        \includegraphics[height=3.5cm]{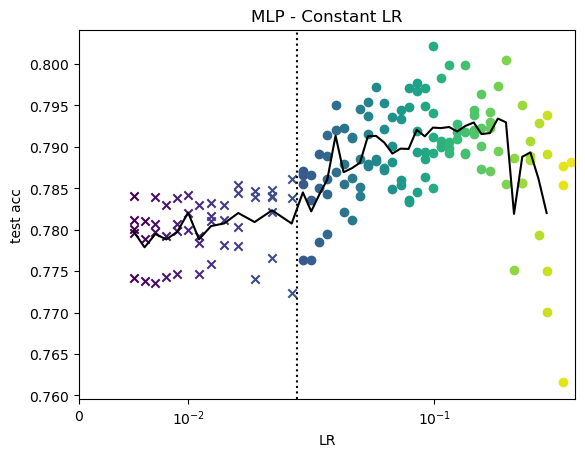}
        \caption{MLP on fMNIST}
    \end{subfigure}%
    ~ 
        \centering
    \begin{subfigure}[t]{0.45\textwidth}
        \centering
        \includegraphics[height=3.5cm]{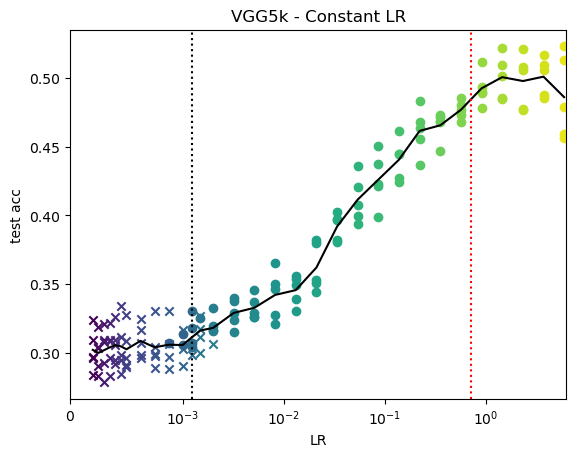}
        \caption{VGG10 on CIFAR10-5k}
    \end{subfigure}%
    ~     
\caption{\textbf{Generalization performance improves past the stability limit.} We train models until completion and plot validation accuracy against the learning rate $\eta_0$. The X/O markers differentiate $\eta_0$s below/above the stability limit (dotted line), and the color spectrum from dark purple to light yellow marks the different learning rates from low to high. }
\label{fig:goldilocks}
\end{figure}  

We study the effects of learning rates ($\eta_0$s) on generalization performance training MLPs on fMNIST and small VGG10s on CIFAR10-5k. To cover a broad range, $\eta_0$s are sampled on an exponential scale starting with $\eta= [0.01, 0.0005]$ (leading to stable trajectories) and scaling factors $m=[1.1, 1.6]$ for each task, respectively. Sampling continues until models diverge, and each model is tested over $5$ random initializations, resulting in a total of $[235,120]$ models for each task. 

Validation accuracy across learning rates are shown in Figure \ref{fig:goldilocks}. For both tasks, the mean accuracy remains relatively flat until $\eta$ goes past the stability threshold, where it sharply improves. This shift highlights the immediate impact of instabilities, which provide notable generalization benefits, as described in earlier sections. Performance eventually falls, indicating that excessive learning rates can be detrimental. These results suggest a \emph{Goldilocks} zone for learning rates. 

The impact on generalization varies between datasets. As CIFAR10 is considered relatively more challenging, regularization plays a more critical role. For this dataset, the regime in which generalization can benefit from large learning rates is large, before the \emph{extremely large regime} regime indicated roughly by the red dotted line. On the other hand,  the improvements on fMNIST is comparatively limited, but the presence of improvements indicate that the regularization effect can be found even with simpler, nearly linearly separable problems.

\subsection{Learning Rate Reduction}\label{sec:gen:exp2-reduc}
\begin{figure}[h]
    \centering
    \begin{subfigure}[t]{0.45\textwidth}
        \centering
        \includegraphics[height=3.5cm]{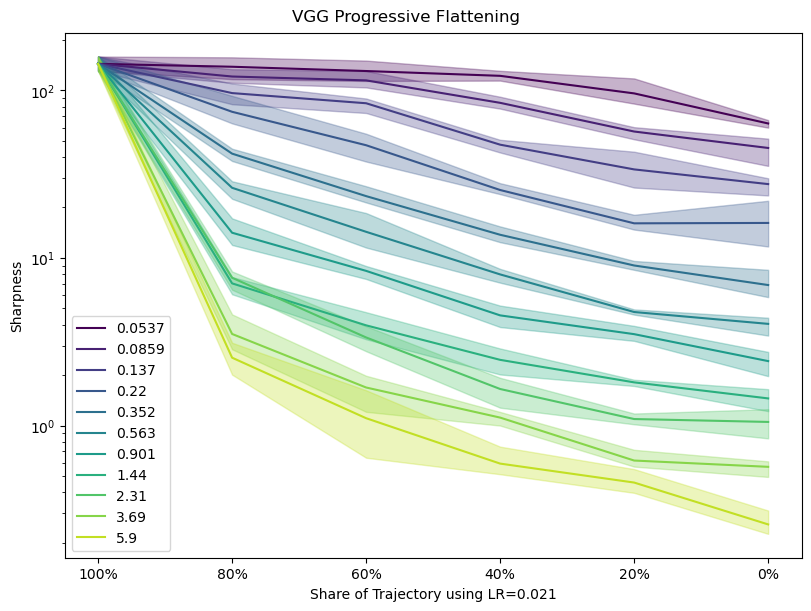}
        \caption{Progressive Flattening}
    \end{subfigure}%
    ~ 
    \centering
    \begin{subfigure}[t]{0.45\textwidth}
        \centering
        \includegraphics[height=3.5cm]{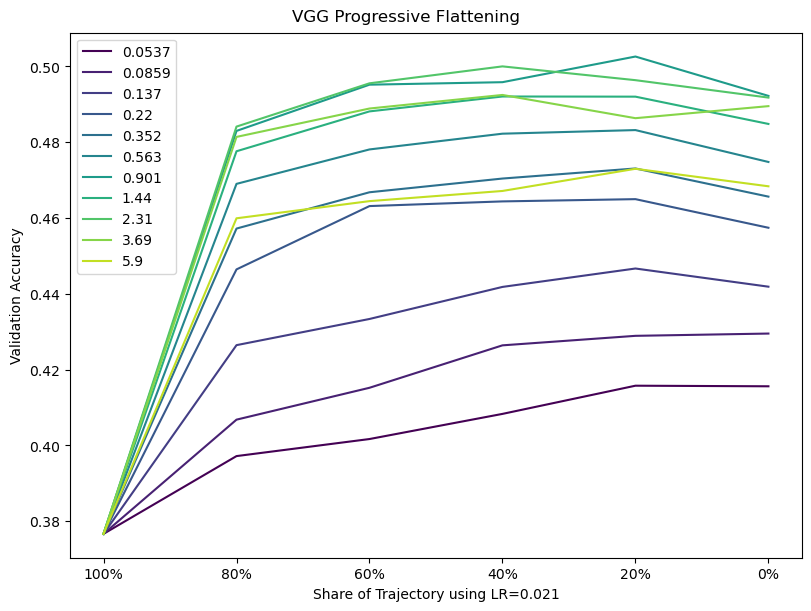}
        \caption{Generalization Performance}
    \end{subfigure}%
    ~ 
    
\caption{\textbf{Training with larger learning rates for longer leads to more progressive flattening and improved generalization.} We train small VGG10s on CIFAR10-5k with large initial learning rates, switching to $\eta_\mathrm{small}=0.021$ at various points of training, until completion. }
\label{fig:cifar-flat}
\end{figure}   

As discussed in Section \ref{sec:rot:emp}, reducing learning rates removes the limit on sharpness to reveal the degree of \emph{progressive flattening}. We begin training small VGG10s with large learning rates $\eta \geq 0.086$, which we later reduce to $\eta_\mathrm{small}=0.021$. The final sharpness and validation accuracies are plotted in Figure \ref{fig:cifar-flat}. As large initial learning rates are applied for increasing amounts of time, the at-completion sharpness of models are reduced, which is indicative of \textit{progressive flattening}. For the majority of models, being flatter improves generalization performance. Additionally, the marginal efficiency of large learning rates diminish, but they remain positive for most models. 

With \textit{extremely large} learning rates, switching to a lower $\eta$ for the final $20\%$ of training (from $\sim 90\%$ accuracy) may improve generalization. This approach is consistent with the popular choice of learning rate reduction towards the end of training, but the benefits of this strategy can be sensitive to timing. 

\subsection{Benchmark on the full Dataset} \label{sec:gen:full-cifar}

\begin{figure}[b]
    \centering
    \begin{tabular}[b]{c|c|c|c|c|c|c}
    $\eta$ & 0.1 & 0.2 & 0.4 & 0.8 & 1.6 & 3.2 \\
    \hline
    \hline
    VGG19 on CIFAR10-50k & 67.09 & 68.36 & 70.53 & 73.41 & 72.70 & 73.56\\
    VGG19 on CIFAR10-500k & 78.12 & 80.63 & 81.73 & 82.88 & 84.01 & 84.07\\
    ResNet20 on CIFAR10-500k & 83.66 & 85.62 & 86.06 & 86.53 & 86.95 & 87.32\\
    \hline
    ResNet20 on CIFAR10-500k; $\eta_\mathrm{small}$ at $90\%$ acc. & n.a. & n.a. & 83.01 & 83.52 & 84.56 & 84.98\\
    ResNet20 on CIFAR10-500k; $\eta_\mathrm{small}$ at $98\%$ acc. & n.a. & n.a. & 85.57 & 85.90 & 86.66 & 86.92\\

    \end{tabular}
    \caption{\textbf{Larger learning rates improve generalization on the unconstrained CIFAR10 datasets.} Learning rate  reduction with ResNet20s uses $\eta_\mathrm{small}=0.021$ until completion. }
\label{fig:cifar50k}

\end{figure}  

Section \ref{sec:gen:exp1-const} suggested a \emph{Goldilocks} zone for $\eta$ and Section \ref{sec:gen:exp2-reduc} suggested that some $\eta$-reduction toward the end of can be beneficial toward generalization. We benchmark these suggestions on the full CIFAR10 dataset, removing constraints imposed on earlier experiments. First, we train small VGG19s on CIFAR10-50k without data augmentations. Next, we introduce non-stochastic data augmentations (crops and flips) by constructing a statically sampled, $10$x augmented CIFAR10-500k dataset, which has shown promising results with prior studies \citep{geiping2022stochastictrainingnecessarygeneralization}. Finally, we evaluate ResNet20s \citep{he2015deepresiduallearningimage} on the augmented CIFAR10-500k dataset, with the resulting validation accuracies across various $\eta$s shown in Figure \ref{fig:cifar50k}. 

The evidence suggests improved generalization with large learning rates. For these models, the stability limits for $\eta$ were not computed due to computational constraints, but the chosen values of $\eta$s were intentionally large to ensure instabilities, as more complex datasets are typically sharper. At $\eta=6.4$, we observed model divergence, which creates a notable gap between this value and the last functioning learning rate, $\eta=3.2$, within which the optimal $\eta$ resides. This highlights a limitation of the exponential sampling method used for learning rates. Nevertheless, our results suggest that the \emph{Goldilocks} zone for $\eta$ lies much closer to the divergence boundary than the stability limit - typically within one order of magnitude of the former and several orders of magnitude from the latter. Consequently, in practice, we recommend using learning rates much higher than what is derived from the \emph{descent lemma}. 

Finally, we explored learning rate reduction in ResNet20s trained on CIFAR10-500k, switching to $\eta_\mathrm{low}=0.1$ at $90\%$ and $98\%$ training accuracies. The observed decline in performance suggests that finding optimal timing for $\eta$ reductions can be challenging in practice. 
 

\section{Related Work}\label{sec:related}


Elements of progressive flattening are observed in the literature. \cite{keskar2017largebatchtrainingdeeplearning} and \cite{jastrzębski2018factorsinfluencingminimasgd} found that training with large learning rates provided benefits toward generalization, which was explained through reductions in sharpness. Our work supports these claims, and we additionally identify a lower bound (only instability-inducing learning rates) where this effect is observed. Moreover, we found that varying the duration of instability can result in different degrees of flattening, aligning with \cite{andriushchenko2023sgdlargestepsizes}'s observation of \emph{loss stabilization}, where phases of large initial learning rates are shown to promote generalization. 

Mechanisms toward the resolution of instabilities have also been studied. \cite{lewkowycz2020largelearningratephase} identified a regime of learning rates, above the stability limit, that can \emph{catapult} models into flatter regions. We identify the potential role of rotations in the catapult effect. \cite{damian2023selfstabilizationimplicitbiasgradient} showed that, considering one unstable eigenvalue, gradient descent has a tendency to self stabilize due to the cubic term in the local Taylor expansion on the constrained trajectory. While acknowledging the importance of higher-order terms, the effects we identify are present without requiring the sharpening factor $\alpha>0$ in our empirical studies (see Appendix \ref{app:drivers}). \cite{arora2022understandinggradientdescentedge} demonstrated that gradient descent can lead to alignment between gradient the sharpest eigenvector of the Hessian $\vv_1$. Our model predicts, and we observe, this alignment during stable phases of training when eigenvectors are reinforced, but during instabilities, the orientations of $\vv_1$ are disrupted. 

Our empirical work contributes to the growing body of research toward assessing sharpness as a metric for generalization. While the lack of scale-invariance is a weakness for sharpness \cite{dinh2017sharpminimageneralizedeep}, remedies have been proposed (e.g. \citep{kwon2021asamadaptivesharpnessawareminimization}). \cite{kaur2023maximumhessianeigenvaluegeneralization} suggest that the pathological nature of these transforms may not arise with standard optimizers. However, recent results by \cite{andriushchenko2023modernlookrelationshipsharpness} on modern benchmarks further challenges the sharpness-generalization link, suggesting that the right measure of sharpness may depend on features of the dataset. Our work (see App \ref{app:gen:exp3-causal}) suggests that sharpness may not be causal to generalization, and we show with preliminary evidence that once learning rates are controlled and differentiated through weight initialization, sharpness can fail to discriminate between models. Given the crucial role of Hessian eigenvectors rotations in resolving instabilities, we encourage further exploration of their role in the dynamics of gradient descent. 


Lastly, \cite{geiping2022stochastictrainingnecessarygeneralization} demonstrated that state-of-the-art performance on CIFAR10 can be achieved using full-batch gradient descent. Our results, also in a non-stochastic setting, replicate their performance (without explicit regularization), using only learning rates to induce regularization via gradient descent instabilities. 

\section{Conclusion}\label{sec:conclusion}

Our work highlights a significant implicit bias in gradient descent via sufficiently large learning rates that favors flatter minima, an effect frequently conjectured in the deep learning literature. We identify rotations of the Hessian eigenvectors as the primary mechanism driving such exploration, which can lead to \emph{progressive flattening} of the loss landscape. Using learning rates above the stability threshold, deep neural networks can realize substantial benefits in generalization, reinforcing the already prevalent use of large learning rates among practitioners. Our experiments are conducted in a fully non-stochastic setting, and encourage further work to determine whether these effects can extend to stochastic gradient descent. We hope our work inspires future efforts toward addressing the role of rotations of the Hessian eigenvectors for a better understanding of optimization with gradient descent.




\bibliography{iclr2025_conference}
\bibliographystyle{iclr2025_conference}







\newpage
\appendix

\section{Justification for the DLN Model in studying rotations}

\subsection{Neural Networks as a Summation of Multiplications}\label{app:dln-reality}

Recapping some notation. In general, let the outputs of a multilayer perceptron (MLP) be:

\begin{align}
    f(\vx) &= \mW_d \sigma_{d-1} (\mW_{d-1} \sigma_{d-2} ( ... \sigma_1(\mW_1 \vx + \vb_1) ... ) + \vb_{d-1} ) +  \vb_d \nonumber
    \intertext{where the $p$-dimensional inputs to the neural network $\vx \in \mathbb{R}^{h_0} = \mathbb{R}^{p}$. Each hidden-layer of index $i, i \in 1, 2, ... d$ has input and output dimensions $\mathbb{R}^{h_{i-1}} \rightarrow \mathbb{R}^{h_{i}}$, weights $\mW_i\in \mathbb{R}^{h_i \times h_{i-1}}$, and bias terms $\vb_i \in \mathbb{R}^{h_i}$. These layers are followed by the activation functions $\sigma_{i}$. For simplicity, we assume a uniform activation function $\sigma_i=\sigma$ to apply to the different layers of the MLP though that may not necessarily be the case, for example with instantiations of modern networks.}
    \intertext{Without loss of generality, let us consider a single-output network such that $h_d = 1$, i.e. a singular output. Additionally, considering the identity activation function, $\sigma=\mI$  (for the moment, we relax this assumption in App. \ref{app:dln-relu}), we get:}
    f(\vx) &= \mW_d \mI(\mW_{d-1} \mI( ... \mI(\mW_1 \vx + b_1) ... ) + \vb_{d-1} ) +  \vb_d  \nonumber \\
    &= (\mW_d \mW_{d-1} ... \mW_1)\vx + \sum^{d}_{j=1} (\mW_{d} \mW_{d-1} ... \mW_{j+1}) \vb_j\nonumber\\
    &= \overbrace{\sum^p_{i=1}\left(\prod_{\textrm{weights along $x_i$'s path}} w_{i, m_d m_{d-1} ...m_1}\right)x_i}^\textit{input contributions} + \nonumber \\
    &\overbrace{\sum^d_{j=1} \left[\sum^{h_j}_{k=1} \left(\prod_\textrm{weights along $b_{j, k}$'s path} w_{k, s_d s_{d-1} ...s_{j+1}}\right)b_{j, k}\right]}^\textit{bias contributions}\label{eqn:sumprod}\\
    \intertext{where $u_{i, m_d m_{d-1} ... m_1}x_i$ represent the scalar at position $i$ in the product $\mW_d \mW_{d-1}...\mW_1$ (recall $h_d=1$), and $u_{k, s_d s_{d-1} ...s_{j+1}}b_{j, k}$ are the contributions $k$th term of the $\mW_d \mW_{d-1}...\mW_{j+1}\vb_j$ product for each term in the summation over $j$. The local path indices $m_d m_{d-1} ...m_1$ and $s_d s_{d-1} ...s_{j+1}$ denote the individual scalar weights along the path of each $x_i$ and $b_{j, k}$, respectively. Specifically: }
    w_{i, m_d m_{d-1} ...q_1}x_i &= \left(w_{\bold{d}, m_d, m_{d-1}} w_{\bold{d-1}, m_{d-1}, m_{d-2}} ... w_{\bold{1}, m_{1}, m_i}\right) x_i \nonumber \\
    w_{k, s_d s_{d-1} ...s_{j+1}} b_{j, k} &= \left(w_{\bold{d}, s_d, s_{d-1}} w_{\bold{d-1}, s_{d-1}, s_{d-2}} ... w_{\bold{j+1}, s_{j+1}, s_k}\right) b_{j, k}\nonumber \\
    \intertext{respectively for summation terms in the input and bias contributions, and recall that $q_d=1$ since $h_d=1$. With the above, we decomposed the matrix formulation of $f(\vx)$ into sum(s) over multiplications of weights. Treating weights and biases both as update-able parameters, each individual term in the bias contributions is equivalent to a DLN parameterization, while each individual term in the input contribution is a DLN scaled by a `constant' $x_i$. In the following sections, we will illustrate the key insights derived from this model in the $2$-parameter DLN, where computations are exact. The proof for general $n$-parameter DLNs is presented in Appendix \ref{app:dln_full}, and the $\sigma=\mI$ assumption is relaxed in Appendix \ref{app:dln-relu} where we consider ReLU activations. } \nonumber
\end{align}

\newpage
\section{Details to the $2$ parameter DLN studied in Section \ref{sec:rot}}\label{app:dln2-workings}
\subsection{From the Loss Hessian to $R_2(\beta)$}\label{app:dln2-beta}
From Section \ref{sec:rot}, we have the loss Hessian:

\begin{align}
\mH(\Theta) &= \begin{bmatrix}
    \frac{\partial^2 L}{\partial \theta_1^2} & \frac{\partial^2 L}{\partial \theta_1 \partial \theta_2} \\
    \frac{\partial^2 L}{\partial \theta_1 \partial \theta_2} & \frac{\partial^2 L}{\partial \theta_2^2} \\
    \end{bmatrix} = \begin{bmatrix}
    z'' \theta_2^2 & z'' \theta_1 \theta_2 +z' \\
    z'' \theta_1 \theta_2 +z' & z'' \theta_1^2 \\
    \end{bmatrix}
\intertext{We solve the characteristic equation in two dimensions, $\mH\vv=\lambda \vv$ to get:}
    \lambda_{1, 2} &= \frac{z''(\theta_1^2+\theta_2^2) \pm \sqrt{z''^2(\theta_1^2-\theta_2^2)^2+4(z'+z''\Theta)^2}}{2} \\
\intertext{Assume for now that $\theta_1>0$ and $\theta_2>0$. We compute the ratio of coordinates:}
    \frac{u_{1, 1}}{u_{1, 2}} &= \frac{\lambda_1 - z''\theta_1^2}{z'+z''\Theta} \notag \\
    &= \frac{z''(\theta_2^2-\theta_1^2) + \sqrt{z''^2(\theta_2^2-\theta_1^2)^2+4(z'+z''\Theta)^2}}{2(z'+z''\Theta)} \nonumber \\ 
\intertext{From the assumptions on $z$ we get $z''>0$. Let us define $r_1=\theta_2^2-\theta_1^2$ and $r_2=\frac{2(z'+z''\Theta)}{z''}$ to get:}
    \frac{u_{1, 1}}{u_{1, 2}} &= g(r_1, r_2)=\frac{r_1 + \sqrt{r_1^2+r_2^2}}{r_2} \nonumber \\
\intertext{When $r_1=0 \longleftrightarrow \theta_1^2=\theta_2^2$, we have $u_{1,1}=u_{1,2}$, which indicates that the sharpest eigenvector is aligned equally to both $\theta$s. The sign of $r_1$ depends on the relative magnitudes of $\theta$s, while the sign of $r_2$ takes the same sign as $\Theta$. We get the conditions:}
    r_1 &> 0, 1 < \left|\frac{u_{1, 1}}{u_{1, 2}}\right| < 1 + \left|\frac{2r_1}{r_2}\right|; r_1 < 0, \left|\frac{u_{1, 1}}{u_{1, 2}}\right| < 1 \nonumber\\
\intertext{which shows the sign-indifference of $\left|\frac{u_{1, 1}}{u_{1, 2}}\right|$ to $r_2$. Since $g(r_1, r_2)g(-r_1, r_2)=1$, we can get the sign-invariant measure of ratios $R_2$:}
    R_2(\beta) &= \beta + \sqrt{\beta^2+1}; \text{ where } \beta=\left|\frac{r_1}{r_2}\right|=\left|\frac{z''(\theta_2^2-\theta_1^2)}{2(z'+z''\Theta)}\right| \label{eqn:rb-def}
\end{align}

\subsection{From gradient updates to $\gamma_\beta$} \label{app:dln2-gamma}
Let $\Delta x$ denote the update $x_{t+1} = x_{t} + \Delta x $, and $\gamma_x$ denote the ratio $\frac{x_{t+1}}{x_t}=\frac{x_t+\Delta x}{x_t}$. 
\begin{align*}
    \Delta \theta_{1} &= \eta \frac{dz}{d\theta_1} = \eta z' \theta_2; \text{ similarly } \Delta \theta_2 =  \eta z' \theta_1 \\
    \gamma_{r_1} &= 1-\eta^2z'^2  \\
    r_2 &= 2\left(\frac{z'}{z''}+\Theta\right) =
    \begin{cases}
    2\Theta ,& \text{when } x \rightarrow 0\\
    c\Theta,              & \text{when } x \rightarrow \infty, \text{where $2 < c \leq 4$ is a constant}
    \end{cases} \\
    \intertext{Given the constant scaling to $\Theta$ in both limits, we approximate the ratio of change $\gamma_{r_2}$ with $\gamma_{r_\Theta}$:}
    \gamma_{r_2} &\approx \gamma_{\Theta} = 1-s\eta z' + \eta^2z'^2, \text{ where } s=\left(\frac{\theta_1}{\theta_2}+\frac{\theta_2}{\theta_1}\right) > 2 \\
    \implies \gamma_\beta &= \frac{\gamma_{r_1}}{\gamma_{r_2}} \approx \frac{\gamma_{r_1}}{\gamma_\Theta} = \frac{1-\eta^2z'^2}{1-s\eta z'+\eta^2z'^2}
\end{align*}

\newpage

\section{Rotations in a General $n$-parameter DLN}\label{app:dln_full}
Let loss $L(\Theta)$ be described by $z(\Theta)$, a non-negative convex polynomial with a unique minimum at $\Theta = 0$, limiting $z(\Theta)$ to even-degree polynomials. Further define:
\begin{align}
    \vv_i &\coloneqq (u_{i, 1}, u_{i, 2}, ..., u_{i, n}) \nonumber \\
    D &\coloneqq z'' \Theta^2 + z' \Theta \nonumber \\
    C_i &\coloneqq \sum^n_j (z'' \Theta^2 + z' \Theta)u_{i,j}\theta^{-1}_j = D \sum^n_j  u_{i,j}\theta^{-1}_j \nonumber \\
    \intertext{We thus get the Hessian:}
    H(\Theta)_{jk} &= (z''\Theta^2 +z'\Theta) \theta^{-1}_j \theta^{-1}_k - z'\Theta \theta^{-1}_j \theta^{-1}_k \delta(j=k) \nonumber \\ 
    &= D \theta^{-1}_j \theta^{-1}_k - z'\Theta \theta^{-1}_j \theta^{-1}_k \delta(j=k) \nonumber \\
    \intertext{As $\mH \vv_i = \lambda_i \vv_i$, so for each coordinate $u_{i, k}$ of $\vv_i$, we have:}
    \lambda_i u_{i,k} &= \left(\sum^n_j Du_{i,j}\theta^{-1}_j\theta^{-1}_k\right) - z'\Theta u_{i,k} \theta^{-2}_k \\
    \left(\lambda_i + \frac{(z'\Theta)}{\theta^2_k}\right)u_{i,k} &= \sum^n_j (z'' \Theta^2 + z' \Theta)u_{i,j}\theta^{-1}_j\theta^{-1}_k = \frac{C_i}{\theta_k}, \nonumber \\
    C_i &= \left(\lambda_i+ \frac{z'\Theta}{\theta_k^2}\right)u_{i,k}\theta_k \label{eqn:C}
\end{align}
    where $C_i$s are constant given $D$. $\lambda_1$ is maximized when $C_1$ is maximized, which we can obtain through constrained optimization:
\begin{align}
    \mathrm{maximize \ }  \frac{C_1}{D} &= \sum^n_j \frac{u_{1,j}}{\theta_j} \label{eqn:optim-c/d} \\
    \mathrm{subject \ to: \ }& \sum^n_j (u_{1,j})^2 - 1 = 0, \textrm{\ \ (\textit{normalization constraint})}\nonumber \\ 
    \intertext{which is maximized by the solution set of $u_1^*$s:}
    \psi :&= u^*_{1,j} \theta_j = \frac{1}{\sqrt{\sum_k^n \theta_k^{-2}}} < \theta_1 \label{eqn:psi} \\
    \sum^n_j \frac{u^*_{1,j}}{\theta_j} &= \psi \sum^n_j \theta_j^{-2} = \psi^{-1} \nonumber \\
\intertext{The set of $u^*_1$s are fixed for each set of $\theta_j$s, and each pair $u^*_{1, j} \theta_j$ is equal to a constant $\psi$. However, this analytical solution is only an approximation because the constancy constraints below were ignored:}
    \mathrm{subject \ to: \ } \forall j, &(\lambda_1 + z'\Theta \theta^{-2}_j) u_{1_j} \theta_j = C_1 = \mathrm{const}, \textrm{\ \ (\textit{constancy constraints})} \nonumber
\intertext{Let $\widehat{u}^*_1$s be the solution set with additional constancy constraints. Clearly, $\sum^n_j \frac{\widehat{u}^*_{1,j}}{\theta_j} \leq \sum^n_j \frac{u^*_{1,j}}{\theta_j}$. Let:} 
    \widehat{u}^*_{1, j} &= u^*_{1, j}(1+\epsilon_{1, j}) \nonumber \\
    \intertext{where the $\epsilon$ are small deviations. Applying the normalization constraint, we get:}
    \sum^n_j &(u^{*, 2}_{1,j})(1+\epsilon_{1, j})^2 - 1 = 0 \label{eqn:norm-constraint} \\
\intertext{Defining $A=\sum^n (u^{*}_{1, j})^2\epsilon$ and $B=\sum^n (u^{*}_{1, j})^2\epsilon^2$, we substitute into \ref{eqn:norm-constraint} to get:}
    \sum^n_j u^{*, 2}_{1, j} + 2A + B - 1 & = 0 \rightarrow G = -2A \nonumber \\
    \textrm{Using \emph{Cauchy-Schwarz}: \ }A^2 &\leq (\sum^n_j (u^{*}_{1, j})^2) B = B \rightarrow  -2 \leq A \leq 0 \nonumber \\
    \textrm{Using Eqn. \ref{eqn:psi} :\ }\sum^n_j \frac{u^*_{1, j}(1+\epsilon_{1, j})}{\theta_j} &= \psi \sum^n_j \frac{1+\epsilon_{1, j}}{\theta_j^{-2}} = (1+A)\psi^{-1} \nonumber
\end{align}
We can hence rewrite the constancy constraints as:
\begin{align}
    \forall j, (\widehat{\lambda}_1+\frac{z'\Theta}{\theta_j^2})(1+\epsilon_{1, j})u^*_{1, j} \theta_j &= D(1+A)\psi^{-1}, -2 \leq A \leq 0 \label{eqn:additional-constraints}    
    \intertext{Where we use $\widehat{\lambda}_1$ to denote values of sharpness that may be different from $\lambda_1$, which are obtained from $u^*_{1, j}$ only using the normalization constraint. Empirically, we find that $A \rightarrow 0^-$ ($A$ approaching $0$ from the negative side), which is consistent with the maximization setting of the problem and that these additional constancy constraints are mild. We now assume that the parameters of the network are ill-conditioned, i.e. $\exists m:  \theta_m^2 \gg \theta_1^2$, as is commonly observed with deep neural networks \citep{papyan2019measurementsthreelevelhierarchicalstructure} and \citep{granziol2021learningratesfunctionbatch}. Using Eqn. \ref{eqn:additional-constraints} we get:}
    j = 1, \ \left(\widehat{\lambda}_1+\frac{z'\Theta}{\theta_1^2}\right) & (1+\epsilon_{1, 1})u^*_{1, 1} \theta_1 = D(1+A)\psi^{-1} \nonumber \\
    j = m: \theta_m^2 \gg \theta_1^2, \left(\widehat{\lambda}_1+\frac{z'\Theta}{\theta_m^2}\right) & (1+\epsilon_{1, m})u^*_{1, m} \theta_m = D(1+A)\psi^{-1} \nonumber\\
    \rightarrow \left(\widehat{\lambda}_1+\frac{z'\Theta}{\theta_1^2}\right)(1+\epsilon_{1, 1})u^*_{1, 1} \theta_1 &= \left(\widehat{\lambda}_1+\frac{z'\Theta}{\theta_m^2}\right)(1+\epsilon_{1, m})u^*_{1, m} \theta_m \label{eqn:epsilon-to-z}\\
    \rightarrow \left(\widehat{\lambda}_1+\frac{z'\Theta}{\theta_1^2}\right) &\approx \widehat{\lambda}_1(1+\epsilon_{1, m}) \nonumber \\
    \textrm{Considering again the $j=m$ case, \ } \widehat{\lambda}_1 (1+\epsilon_{1, m})\psi &= D(1+A)\psi^{-1} \approx D\psi^{-1} = \lambda_1 \psi \nonumber \\
    \rightarrow \widehat{\lambda}_1(1+\epsilon_{1, m}) & = \lambda_1 \nonumber \\
    \textrm{substituting into Eqn \ref{eqn:epsilon-to-z}: \ } \epsilon_{1, m} \approx \frac{z'\Theta}{\theta_1^2 \widehat{\lambda}_1} & =\frac{z'\Theta(1+\epsilon_{1, m})}{\theta_1^2 \lambda_1} = \frac{z'\Theta(1+\epsilon_{1, m})}{\theta_1^2 D \psi^{-2}}\nonumber \\
    0 < \epsilon_{1, m} &< \frac{z'\Theta}{z''\Theta^2} \nonumber
    \intertext{where $0<\frac{z'\Theta}{z''\Theta^2}\leq 1$, where the right-hand equality holds for quadratic $z$. Note that this implies:}
    \lambda_1  > \widehat{\lambda}_1 > & 0, \nonumber \\
    \widehat{\lambda}_1 > \widehat{\lambda}_1\epsilon_{1, m} \approx & \frac{z'\Theta}{\theta_1^2} \nonumber
    \intertext{Applying the constancy constraints we can get a ratio of parameters of the sharpest eigenvector, $\vv_1$:}
    R_n(k) = \left|\frac{u_{1,1}}{u_{1,k>1}}\right| &= \frac{\left|\widehat{\lambda}_1\theta_k + z'\Theta/\theta_k\right|}{\left|\widehat{\lambda}_1\theta_1 + z'\Theta/\theta_1\right|} \nonumber \\
    &= \frac{|\theta_k|}{|\theta_1|} \frac{\widehat{\lambda}_1 + z'\Theta/\theta_k^2}{\widehat{\lambda}_1 + z'\Theta/\theta_1^2} = \frac{f(\theta_k)}{f(\theta_1)} \label{eqn:R-general} \\ 
    \intertext{Since $\widehat{\lambda}_1>0$ and $z'\Theta>0$, the function $f(\theta) \coloneq |\theta| (\widehat{\lambda}_1 + z'\Theta/\theta^2)$ is positive and has positive derivatives when $|\theta|>\theta_\mathrm{crit}=(\widehat{\lambda}_1/z'\Theta)^{-0.5}$. The bound $\widehat{\lambda}_1>z'\Theta/\theta_1^2$ implies $\forall j: |\theta_j| > \theta_\mathrm{crit}$, which makes this a monotonically increasing function in $|\theta|$ for our domain, implying $R_n \geq 1$. Additionally, write: }
    \forall k: \theta_k^2 = r_k^2\theta_1^2 \rightarrow \Theta &= \theta_1^n \prod^n_k |r_k|, r_k^2 \geq 1 \nonumber \\
    \intertext{where $r_k^2$ is defined as a ratio of relative parameter magnitudes. By the scale-invariance of eigenvector orientations (i.e. multiplying each $\theta$ by a constant should not reorient the eigenvectors)\footnote{This is also empirically verified.}, we have:} 
    \frac{dR_n(k)}{d\theta_1} &= 0 \label{eqn:sym-eig} \nonumber
    \intertext{but:}
    \frac{dR_n(k)}{dr_k} &= \frac{\theta_1 f'(r_k\theta_1)}{f(\theta_1)}>0 \nonumber
\end{align}

The interpretation of $R_n$ is similar to $R_2(\beta)$ from the $n=2$ case (Eqn. \ref{eqn:rb-def}) - a large $R_n$ indicates strong alignment to the unstable parameter, and vice versa. With the above results, we find that $R_n$ yields remarkably similar behavior to $R_2(\beta)$. Define (relatively) sharp and flat parameters $\theta_s, \theta_f:\theta_f^2/\theta_s^2 = r_{f/s} > 1$, we characterize the phases of learning:
\begin{enumerate}
    \item In the stable phase of learning, the ratio of parameters $r_{f/s}$ increases, which leads to an increase in $R_n$, This signifies increased alignment of the sharpest eigenvector $\vv_1$ to $\theta_1$, the unstable parameter. 
    \item During an instability, the ratio of parameters $r_{f/s}$ falls, leading to a decrease in $R_n$. This implies that the $\vv_1$ rotates away from $\theta_1$, the unstable parameter
    \end{enumerate}
A detailed phase-analysis based on $\eta$ can be conducted with a strategy similar to that employed in the $n=2$ case. 
\section{Extension to the ReLU activation function}\label{app:dln-relu}
In Section \ref{app:dln_full}, we derived the orientation of eigenvectors:
\begin{equation}
    \frac{dR_n(k)}{dr_k} = \frac{xf'(r_k\theta_1)f(\theta_1)}{f(\theta_1)^2} > 0 \nonumber
\end{equation}
where $R_n(k)$ was a positive scalar indicating the degree of alignment of $\vv_1$ to the unstable parameter, and $r_k$ was the ratio of the parameter magnitudes. In this section, we relax the $\sigma=\mI$ assumption and provide extensions of these results to ReLU activations.

Our claims are relaxed to the $2$-step process under gradient descent, where weaker assumptions on the form of $z$ can be made to accommodate nonlinear activations functions. Specifically, we only assume that $z$ has a unique minimum at $\Theta=0$ (without loss of generality due to translation invariance), and is continuous and convex for each side of $\Theta \in \mathbb{R}^+, \mathbb{R}^-$ but, not necessarily at the minimum. This means that the function is not necessarily symmetric. 

Since the proof in Appendix \ref{app:dln_full} only uses the sign-properties of $z$ through $z\Theta>0$, which also holds true under this relaxation, we therefore expect the proof to generalize when considering only 'half' of the function ($\Theta>0$ or $\Theta<0$), motivating our strategy to study the period $2$ process. 

To demonstrate the existence of stability, we begin by assuming that for each $\Theta$, there exist large and small learning rates, $\eta_l, \eta_s$, such that they lead to an increase or decrease in $\Theta$ after two updates:
\begin{align}
    \forall \Theta_t \neq 0, &\exists \eta_{l}: \Theta_{t+2} = \Theta_{t+1} - \eta_{l} z'(\Theta_{t+1}) = \Theta_{t} - \eta_{l} z'(\Theta_{t}) - \eta_{l} z'(\Theta_{t+1}) > \Theta_t \nonumber\\
    \forall \Theta_t \neq 0, &\exists \eta_{s}: \Theta_{t+2} = \Theta_{t+1} - \eta_{s} z'(\Theta_{t+1}) = \Theta_{t} - \eta_{s} z'(\Theta_{t}) - \eta_{s} z'(\Theta_{t+1}) < \Theta_t \nonumber\\
    \intertext{which reflects mild assumptions that closely match widespread empirical observations. When the minimum $\Theta$ is not reached, $\Theta_{t+2}$ depends continuously on $\eta$. We apply the \textit{Intermediate Value Theorem} for the existence of an equilibrium, and the corresponding learning rate $\eta_\mathrm{eq}$:}
    \forall \Theta_t, &\exists \eta_\mathrm{eq}: \Theta_{t+2} = \Theta_{t+1} - \eta_\mathrm{eq} z'(\Theta_{t+1}) = \Theta_{t} - \eta_\mathrm{eq} z'(\Theta_{t}) - \eta_\mathrm{eq} z'(\Theta_{t+1}) = \Theta_t \nonumber\\
    \intertext{which implies the existence of a two-stage dynamically stable trajectory. We consider a linearization of ReLU $\sigma(x) = \mathrm{max}(0, x)$:}
    f(x) &= \mW_n \sigma_{n-1} (\mW_{n-1} \sigma_{n-2} ( ... \sigma_1(\mW_1 \vx + \vb_1) ... ) + \vb_{n-1} ) +  \vb_n \nonumber \\
    &= \mW_n \mM_{n-1} (\mW_{n-1} \mM_{n-2} ( ... \mM_1(\mW_1 \vx + \vb_1) ... ) + \vb_{n-1} ) +  \vb_n \nonumber
\end{align}
where $\mM_{i} \in \mathbb{R}^{h_i\times h_i}$ is a diagonal masking matrix with elements $[1, 0]$ in its diagonals representing the active/inactive states of the ReLU units at layer $i$. This formulation can be decomposed to the sum of products similar to the strategy employed in Eqn. \ref{eqn:sumprod}, given that $\mM$s are constant. 

Consequently, our characterization of $\frac{dR_n(k)}{dr_k}$ applies to the ReLU MLPs when considering eigenvector rotations in a period of $2$ with respect to the training epochs, up until the limit of the signs of pre-activations are constant, which holds when $(\eta-\eta_\mathrm{eq})$ is small. 

\begin{figure}[b] 
\includegraphics[width=\linewidth]{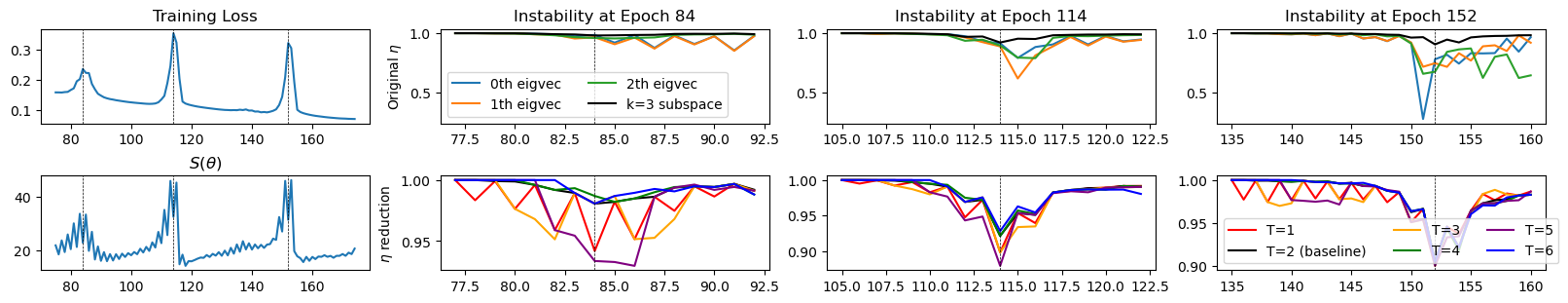}
\caption{\textbf{During instabilities, the sharpest eigenvectors of the Hessian rotate away smoothly and monotonically (Top, same as Fig \ref{fig:instab_rot}. Smooth rotations only occur for even periods $T\%2=0$.} We track the similarity of the sharpest Hessian eigenvectors across epochs through three instabilities. \textbf{Left:} $L(\theta)$ and $S(\theta)$. \textbf{Top:} similarities of the $k$-th eigenvectors (colored) and of subspaces formed by the top $3$ eigenvectors (black) during instabilities with period $T=2$. \textbf{Bottom:} similarity of subspaces formed by the top $3$ eigenvectors with varying period $T$s. }
\label{fig:instab_rot_oddeven}
\end{figure}

With these considerations, we expect our theoretical predictions of eigenvector rotation to apply to ReLU activations at a period of $T=2$. These predictions are corroborated by empirical evidence in Figure \ref{fig:instab_rot}, where the loss landscape for this ReLU MLP is not symmetric, which was visualized using comparison period $T=2$. In \ref{fig:instab_rot_oddeven}, we plot eigenvector similarities for a number of different period $T$s. Gradual rotations in the eigenvectors are observed over even-period processes, and not for adjacent odd-period processes. We expect a similar decomposition strategy to apply to other nonlinear activation functions, suggesting the suitability of the DLN model as a general tool for analysis of individual terms in the loss function of deep neural networks. 

\newpage
\section{Extension to general outputs that are sums of DLNs (multiplicative terms)}\label{app:dln-loss}

We motivated the use of a DLN as a model with the SoM decomposition of the outputs $f(\vx)$ of an MLP. We then studied the rotational behaviors of the DLN through $R_n$, the ratio of parameters, to draw insights and claims on the rotational behavior of Hessiain eigenvectors. In previous sections, we studied a polynomial function $z$ as an approximation to the loss function, through a single DLN (multiplicative term) through which the summation represents general MLP outputs. Here, we offer strategies to generalize our previous results to an arbitrary sum of DLN terms, providing justification of our insights on eigenvector rotation, which were derived from a simple DLN model, to general formulations of MLPs. The extension the ReLU activation is studied earlier in Appendix \ref{app:dln-relu}. We study the $2$-term addition model first, before offering a strategy for general $m$-term additive models. 

\subsection{Two-term additive case}
We consider the loss function that is approximated by the addition of two DLN terms, $L(\Theta_a, \Theta_b) \approx z(\Theta_a, \Theta_b) = (\Theta_a + \Theta_b)^q$, where $q$ is an even integer, following our assumptions about $z$ in Section \ref{sec:rot:dln2}. Each DLN output is composed as $\Theta_a = \prod^m_i=1 (\theta_i)$ and $\Theta_b=\prod^{n+m}_{i=1} (\theta_i+m)$. Write $z':=\frac{\partial z}{\partial \Theta_a}$ and $z^\crosssymbol:=\frac{\partial z}{\partial \Theta_b}$, we can write the combined Hessian $\mH(\Theta_a, \Theta_b)$: 

\begin{align}
    \mH &= \left[\begin{array}{c c c : c c c}
        z'' \Theta_a^2 \theta_1^{-2} & ... &  z'' \Theta_a^2 \theta_1^{-1}\theta_m^{-1} & z^{\prime\crosssymbol} \Theta_a \Theta_b \theta_1^{-1}\theta_{m+1}^{-1}& ...& z^{\prime\crosssymbol} \Theta_a \Theta_b \theta_1^{-1}\theta_{n}^{-1}\\
        \vdots & & \vdots & \vdots & & \vdots \\
        z'' \Theta_a^2 \theta_m^{-1}\theta_1^{-1} & ... &  z'' \Theta_a^2 \theta_m^{-2} & z^{\prime\crosssymbol} \Theta_a \Theta_b \theta_m^{-1}\theta_{m+1}^{-1}& ...& z^{\prime\crosssymbol} \Theta_a \Theta_b \theta_m^{-1}\theta_{n}^{-1}\\[4pt] \hdashline
        \\[-8pt]
        z^{\crosssymbol\prime} \Theta_b \Theta_a  \theta_{m+1}^{-1} \theta_1^{-1} & ... &  z^{\crosssymbol\prime} \Theta_b \Theta_a  \theta_{m+1}^{-1} \theta_m^{-1} & z^{\crosssymbol\crosssymbol} \Theta_b^2 \theta_{m+1}^{-2}& ...& z^{\crosssymbol\crosssymbol} \Theta_b^2 \theta_{m+1}^{-1} \theta_{n}^{-1}\\
        \vdots & & \vdots & \vdots & & \vdots \\
        z^{\crosssymbol\prime} \Theta_b \Theta_a  \theta_{n}^{-1} \theta_1^{-1} & ... &  z^{\crosssymbol\prime} \Theta_b \Theta_a  \theta_{n}^{-1} \theta_m^{-1} & z^{\crosssymbol\crosssymbol} \Theta_b^2 \theta_n^{-1} \theta_{m+1}^{-1}& ...& z^{\crosssymbol\crosssymbol} \Theta_b^2 \theta_{n}^{-2}\\
        \end{array}\right]  \nonumber\\
        &= \begin{bmatrix}
        \mH(\Theta_a) & \mG(\Theta_a, \Theta_b)  \\
        \mG(\Theta_b, \Theta_a) & \mH(\Theta_a) \\
        \end{bmatrix} \label{eqn:Hess-block-form-2}
\end{align}

where dependencies on $\Theta_a, \Theta_b$ have been dropped for $z$ and $\mH$. The combined Hessian $\mH$ can be decomposes into individual-DLN Hessians $\mH(\Theta_a)$ and $\mH(\Theta_b)$ on the diagonals, and the cross-term Hessians $\mG(\Theta_a, \Theta_b)$, $\mG(\Theta_b, \Theta_a)$. When scales of the DLN outputs are imbalanced, e.g. $\Theta_a^2 \gg \Theta_b^2$, the eigenvalues and eigenvectors of the combined Hessian can be approximated by studying the individual-DLN Hessian, $\mH(\Theta_a)$. When the DLN outputs are similar in scale, the analysis is complicated by the non-negligible contributions from the cross-term Hessians, $\mG$.

Following a similar strategy to Appendix \ref{app:dln_full}, we use constrained optimization on the characteristic equation to solve for $u$s. Recall $u_{i, j}$ is the component of the $i$-th eigenvector in the basis along parameter $\theta_j$. Define:
\begin{align}
    D_a&\coloneq z''\Theta_a^2+z'\Theta_a, D_b \coloneq z^{\crosssymbol\crosssymbol}\Theta_b^2+z^{\crosssymbol}\Theta_b, D_{ab} = D_{ba} \coloneq z^{\prime\crosssymbol}\Theta_a\Theta_b \nonumber\\
    \intertext{using the characteristic equation, $\mH\vv = \lambda_i \vv$:}
    \lambda_i u_{i, k} &= \sum_{j}^n H_{k,j} u_{i, j} \nonumber \\
    &= \begin{cases}
        \sum_{j}^n  \left( \overbrace{D_a \delta(j \leq m) - z' \Theta_a \delta(j=k)}^{\textit{$\mH(\Theta_a)$ contributions}} +\overbrace{D_{ab} \delta(j > m))}^{\textit{$\mG$ contributions}} \right) \theta_j^{-1} \theta_k^{-1} u_{i, j} , k \leq m \\
        \sum_{j}^n \left( \overbrace{D_b \delta(j > m) - z^{\crosssymbol} \Theta_b \delta(j=k)}^{\textit{$\mH(\Theta_b)$ contributions}} +\overbrace{D_{ab} \delta(j \leq m))}^{\textit{$\mG$ contributions}} \right) \theta_j^{-1} \theta_k^{-1} u_{i, j}, k > m \\
    \end{cases} \nonumber
    \intertext{additionally define $E_i$:}
    E & \coloneq \delta(k\leq m) \left[D_a\delta(j\leq m) + D_{ab} \delta(j>m)\right] + \delta(k > m) \left[D_b\delta(j > m) + D_{ab} \delta(j\leq m)\right] \nonumber \\
    \intertext{we get the constancy conditions, as in Eqn \ref{eqn:C}:}
    C_i &= \sum_{j}^n E \theta_j^{-1} u_{i, j} = \left(\lambda_i + \frac{z'\Theta_a\delta(k \leq m) + z^\crosssymbol \Theta_b\delta(k > m)}{\theta_k^2} \right)\theta_k u_{i, k} \nonumber
\end{align}
where $C_i$s are constant given $E$. We apply constrained optimization to maximize $C_1/E$ with the normalization constraint to get the set of $u_{1}$s:
\begin{align}
    \textrm{maximize \ } & C_1/E =  \sum_{j}^n \theta_j^{-1} u_{1, j} \nonumber \\
    \textrm{subject to: }& \sum_j^n (u_{1, j}^2) - 1 = 0, \emph{normalization constraints} \nonumber \\
    \textrm{optimal solutions $u^*_1$s fulfill: } & \psi = u_{1, j}^* \theta_j = \frac{1}{\sqrt{\sum_k^n \theta_k^{-2}}} <  \theta_\mathrm{min} \nonumber
\end{align}
which takes the same maximum $\psi^{-1}$, using the same  derivation as Equation \ref{eqn:optim-c/d}. The application of the constancy constraints are more complex in this case. Incorporating the deviation as in \ref{eqn:additional-constraints}, we have:
\begin{align}
    \forall j, \left(\widehat{\lambda}_1+\frac{z'\Theta_a\delta(j\leq m) + z^{\crosssymbol}\Theta_b\delta(j > m)}{\theta_j^2}\right)(1+\epsilon_{1, j})u^*_{1, j} \theta_j &= E(1+A)\psi^{-1}, -2 \leq A \leq 0 \nonumber 
\end{align}

where $A$ is defined similarly as $A = \sum^n (u^*_{1, j})^2\epsilon$. Consider two indices for comparison, $i, j$, and let $\theta_i$ be the smaller parameter i.e. $\theta_i^2 \ll \theta_j^2=\theta_i^2 r_{ij}^2$. When both indices are in the same regime ($i, j \leq m$ or $i, j > m$), the computation of $R_n$ follows the same reasoning as in Appendix \ref{app:dln_full} since $E$ is constant under these domains. Specifically:

\begin{equation*}
    \begin{cases}
        (\widehat{\lambda}_1+\frac{z'\Theta_a}{\theta_j^2})(1+\epsilon_{1, j})u^*_{1, j} \theta_j = (D_a + D_{ab})(1+A)\psi^{-1}, & \forall j \leq m\\ 
        (\widehat{\lambda}_1+\frac{z^{\crosssymbol}\Theta_b}{\theta_j^2})(1+\epsilon_{1, j})u^*_{1, j} \theta_j = (D_b+D_{ab})(1+A)\psi^{-1}, & \forall j > m
    \end{cases}    
\end{equation*}

Define:
\begin{align}
    f_y(x) &\coloneq |x|\widehat{\lambda_1} + \frac{\partial z}{\partial y} y / x^2 \nonumber\\
    \intertext{By the constancy of $E$ when regimes of $i, j$ are equal, we have:}
    f{\Theta_a}(\theta_j)&>0 \text{\ and \ } f'_{\Theta_a}(\theta_j)>0 \nonumber\\
    \intertext{We study $R_n$ for when regimes of $i, j$ are separated. Let $i$ be a part of $\Theta_a$ and $j$ a part of $\Theta_b$, so $i \leq m$ and $j > m$ wlog, we have:}
    R_n(i_a, j_b) &= \frac{f_{\Theta_b}(\theta_j)}{f_{\Theta_a}(\theta_i)} \nonumber
    \intertext{By symmetry arguments, we have $\frac{\partial R_n(i_a, j_b)}{\partial \theta_j} = 0$. Additionally:}
    \frac{\partial R_n(i_a, j_b)}{\partial r_{ij}} &= \frac{\theta_i f'_{\Theta_b}(r_{ij}\theta_i)}{f_{\Theta_a}(\theta_i)} >0\nonumber
\end{align}
since $\frac{\partial R_n(i_a, j_b)}{\partial \theta_j} = 0$ and $\frac{\partial R_n}{\partial r_{ij}} > 0$ for all cases of $i, j$, our phase analysis and claims about eigenvector rotation from Appendix \ref{app:dln_full}, the one-DLN $n$-parameter case extends to the two-DLN $n$-parameter additive case. 

\subsection{General strategy for the $m$-term additive case}
In general, the addition of $m$ DLN terms as inputs to the polynomial loss function is $z(\Theta_1, \Theta_1, ..., \Theta_m) = (\Theta_1, \Theta_2, ..., \Theta_m)^q$, where $q$ is an even integer. Here, we use $1...m$ as indices for $\Theta$s, for notational clarity, where we used indices $a, b$ in the previous section. This will have a Hessian matrix $\mH(\Theta_1, \Theta_2, ..., \Theta_m)$ that takes a predictable block structure, as in Eqn \ref{eqn:Hess-block-form-2}:

\begin{equation}
    \mH(\Theta_1, \Theta_2, ..., \Theta_m) = \begin{bmatrix}
        \mH(\Theta_1) & \mG(\Theta_1, \Theta_2) & ... & \mG(\Theta_1, \Theta_m) \\ 
        \mG(\Theta_2, \Theta_1) & \mH(\Theta_2) & ... & \mG(\Theta_2, \Theta_m) \\ 
        \vdots & \vdots & &\vdots \\
        \mG(\Theta_m, \Theta_1) & \mG(\Theta_m, \Theta_2) & ... & \mH(\Theta_m) \\ 
    \end{bmatrix}
    \nonumber
\end{equation}
where $\mH(x)$ is the Hessian of $z(x)$, and $\mG(x, y)$ the cross-term Hessians as \ref{eqn:Hess-block-form-2}. 
The strategy is as follows:

\begin{enumerate}
    \item By using the characteristic equation, obtain scaled constant terms separated by DLN attribution, e.g. :
    \begin{align}
        D_i \coloneq \frac{\partial^2 z}{\partial \Theta_i ^ 2}\Theta_i^2 + \frac{\partial z}{\partial \Theta_i}\Theta_i, & \text{\ and \ } D_{ij} \coloneq \frac{\partial^2 z}{\partial \Theta_i \partial \Theta_j}\Theta_i \Theta_j\nonumber\\
        \left(\lambda_i + \frac{\partial z}{\partial \Theta_a}/\theta_k^2\right)\theta_k u_{i, k} &= \left[D_{a1} + D_{a2} + ... + D_a + ... + D_{am} \right]\sum_j^n u_{i, j} \theta_j^{-1} \nonumber
    \end{align}
    where the parameter $\theta_k$ belongs to DLN-term $\Theta_a$, without loss of generality
    \item Treating $\left[D_{a1} + D_{a2} + ... + D_a + ... + D_{am} \right]$ as constants, we use constrained optimization on $\sum_j^n u_{i, j} \theta_j^{-1}$ to derive locally optimal solutions for $\lambda_1$ and $\vv1$, the latter in the form of a specific set of $u^*_1$s, as in the optimization in Eqn \ref{eqn:optim-c/d}
    \item We then incorporate minor deviations from the lack of the constancy constraint (in the form of $(1+A)$), and derive ratio of parameters $R_n(i, j)$. Additionally, we obtain DLN-dependent functions of ratios (for any $\theta_k$ being attributed to $\Theta_a$, wlog):
    \begin{align}
        f_{\Theta_a}(\theta_k)&=|\theta_k|\widehat{\lambda_1} + \frac{\partial z}{\partial \theta_k} /\theta_k^2 \nonumber
    \end{align}
    \item Considering each $\Theta_b$, $b \in [1, 2, ... m]$, we can derive the bound $\widehat{\lambda_1}>\frac{\partial z}{\partial \theta_k} /\theta_k^2$, which implies that $\frac{df_{\Theta_b}}{\theta_{b, \mathrm{min}}}>0$, where $\theta_{b, \mathrm{min}}$ is the smallest $\theta$, in absolute value, attributed to the DLN $\Theta_b$
    \item We define $\theta_j^2 = \theta_i^2 r_{ij}^2$, where we assumed $\theta_j^2>\theta_i^2$ wlog, and write down $R_n(i, j)$ 
    \item We use symmetry arguments for $\frac{dR_n(i, j)}{d \theta_j} = 0$
    \item Finally, we show that regardless of the DLN-attribution of $\theta_i$ and $\theta_j$, we have $\frac{d R_n(i, j)}{d r_{ij}}>0$
\end{enumerate}
The results of the derivatives of $R_n$ with respect to the parameters and the ratio between any set of them allows us to conduct the phase analysis of rotations as we did in Section \ref{app:dln_full}. 
\subsection{Limitations}
While we offer a strategy for general $n$-parameters and $m$ DLNs, it relies on crucial assumptions. Specifically, our derivations require that the deviations of $u_1$ to $u_1^*$, as captured by $\epsilon$ and $A$, are small. We studied this empirically for one DLN loss function and were not able to falsify this claim, though this may not necessarily be the case for for $m$ DLN formulations. Additionally, we assume that the parameters are ill-conditioned within each DLN. This is sensible and well-researched for deep neural networks, e.g. the works of \cite{granziol2021learningratesfunctionbatch} and \cite{papyan2019measurementsthreelevelhierarchicalstructure}, and hence is a reasonable assumption to make across the neural network. However, this may not necessarily hold true for each individual DLN in the $m$-term summation. We leave the verification of these assumptions to future work. 

\newpage
\section{Technical Details of Experiments} \label{app:tech}
\begin{figure}[h] 
\includegraphics[width=\linewidth]{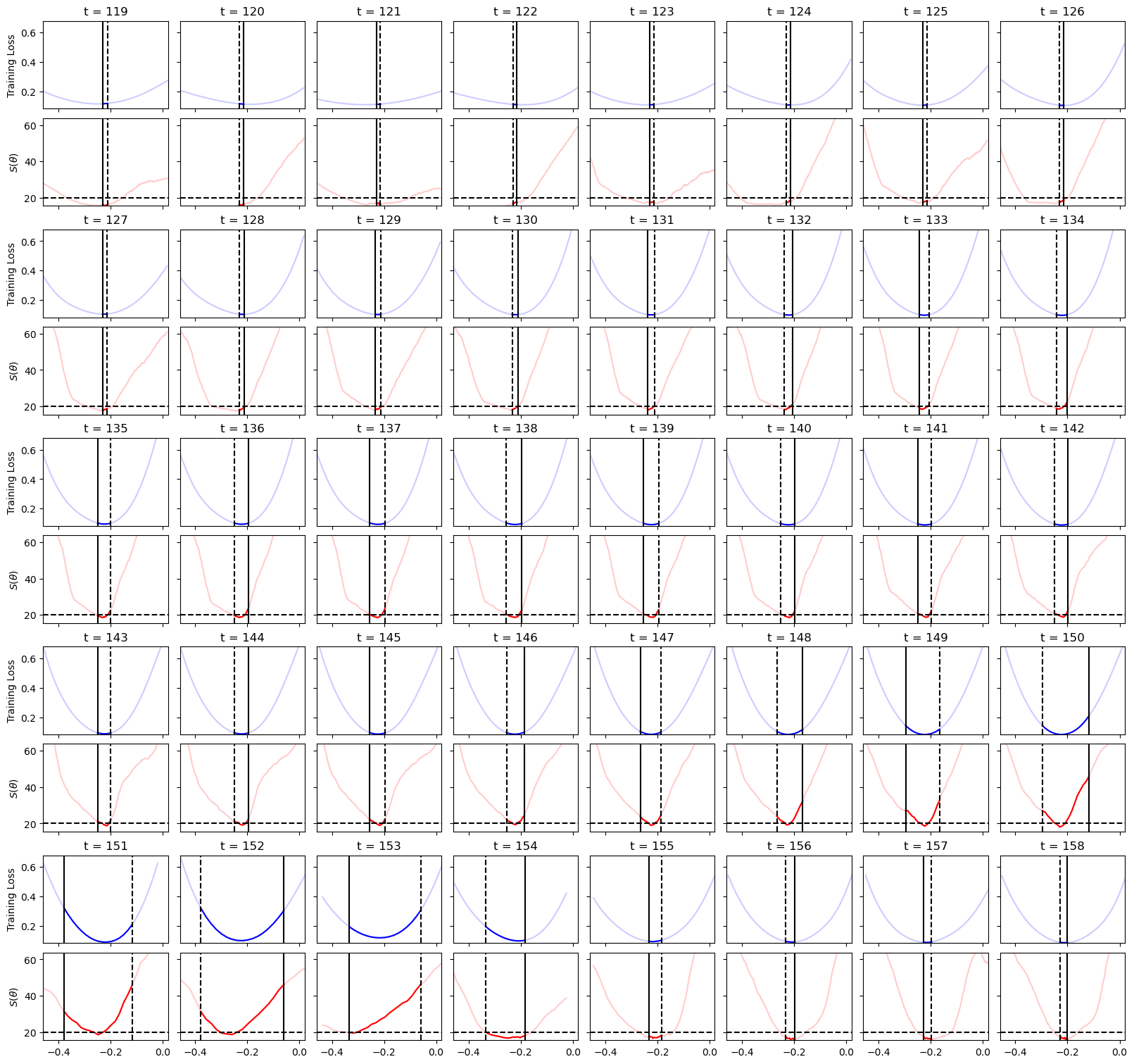}
\caption{\textbf{Parameter growth along the sharpest Hessian eigenvectors leads to exploration of the peripheries of the local minima, driving up $L(\theta)$ and $S(\theta)$ in the process. As the instability develops, the $S(\theta)$ curve undergoes large changes until a flat region is found to enable a return to stability.} We show snapshots along the instability cycle taken along the direction of the gradient. The dotted/solid vertical lines indicate the positions of previous/current parameters, respectively. }
\label{fig:giant-movie}
\end{figure}

A large number of experiments were conducted in this work. All of our models are small, to faciliate ease of computation. These models are trained in a fully non-stochastic setting, using full-batch gradient descent. Moreover, we refrained from the use of common techniques promoting stability, such as momentum and RMSProp \citep{kingma2017adammethodstochasticoptimization}, and explicit regularizers, such as weight decay. We employed the following experimental settings: 

\begin{enumerate}
    \item Figure \ref{fig:cartoon} used a toy $2$-parameter DLN model.
    \item \textbf{MLPs on fMNIST:} The MLPs consisted of $4$ hidden layers, each of width $32$, for a total of $28480$ parameters. This model was trained on $1,000$ samples of fMNIST and evaluated on $200$. The dataset was pre-processed with standard normalization. This setting was used in Figures \ref{fig:instab_rot}, \ref{fig:movie}, \ref{fig:prog-flat}, \ref{fig:goldilocks}a), \ref{fig:corr}a), \ref{fig:phases}, \ref{fig:control-pc}, and \ref{fig:giant-movie}. 
    \item \textbf{Small VGG10s on CIFAR10-5k:} The VGG10s consisted of $3$ VGG blocks with GhostBatchNorm computed at fixed batch size $1,024$ and no dropout. The VGG blocks had $3$ convolutional layers, with each block increasing in width ${8, 16, 32}$, leading to a total of $47,892$ parameters (including BatchNorm params). During training, a fixed $100$ epochs were dedicated to a linear warmup schedule for learning rates. This model was trained on $5,000$ samples of CIFAR10 and evaluated on $1,000$, pre-processed with standard normalization. This setting was used in Figures \ref{fig:goldilocks} b), \ref{fig:cifar-flat}, \ref{fig:corr}b) c), \ref{fig:cifar-kde}. 
    \item \textbf{Small VGG19s and ResNet20s on CIFAR10-50k and CIFAR10-500k:} The VGG19s consisted of $6$ VGG blocks with $3$ convolutional layers, with each block increasing in width ${16, 32, 64, 128, 256, 512}$, leading to a total parameter count of $335,277$. The ResNet follows the standard architecture outlined \cite{he2015deepresiduallearningimage}, where the identity function is used as residual connections as per the original paper, using $271,117$ parameters. Both models used GhostBatchNorm in place of BatchNorm computed at fixed batch size $1,024$ and no dropout. During training, a fixed $100$ epochs were dedicated to a linear warmup schedule for learning rates. These models were trained on $50,000$ and $500,000$ samples of CIFAR10, and evaluated on the full $10,000$ samples available, using the default train-test split and pre-processed with standard normalization. The $500$k dataset uses static data augmentations, where each sample from the original $50$k is used to generate $10$ fixed augmentations, using a combination of random crop and horizontal flips. This setting was used in Figure \ref{fig:cifar50k}. 
\end{enumerate} 
\newpage
\section{Drivers of Instability}\label{app:drivers}

\begin{figure}[h] 
\includegraphics[width=\linewidth]{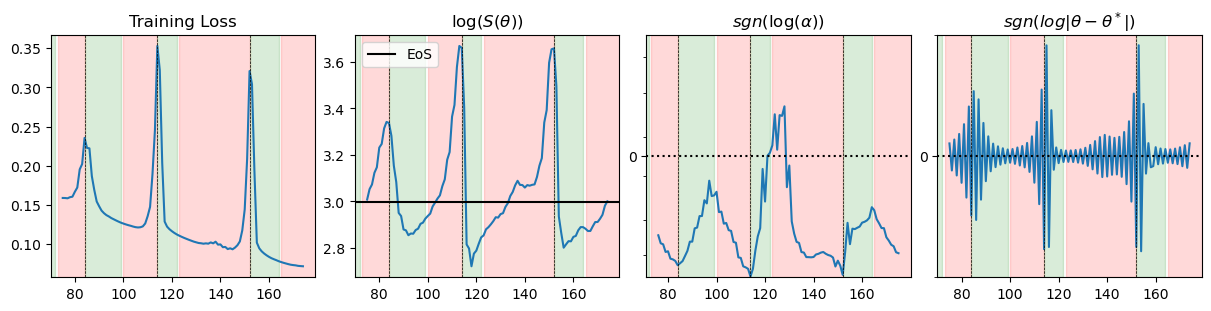}
\caption{\textbf{Magnitude of oscillations divides stable and unstable phases of training.} The turning points of $|\theta - \theta^*|$, distance to mean oscillation parameters, are used to demarcate phases of learning on an MLP trained on fMNIST with FBGD. }
\label{fig:phases}
\end{figure}   

We present an exploration of the training dynamics with gradient descent to pinpoint the sources of instability. The Edge of Stability is reached when $S(\theta)$ rises to $2/\eta$ early in training through progressive sharpening. At this stage, the model exhibits unstable behavior characterized by sudden spikes in training loss and $S(\theta)$, dubbed \emph{instabilities}. To clearly illustrate these dynamics, we train a multi-layer perceptron (MLP) on the fMNIST dataset \citep{xiao2017fashionmnistnovelimagedataset}, a simple model on a straightforward task. 


Classical theories of stability posits that once past the stability limit, oscillations in parameters become unstable, leading to an increase in magnitude to eventually leads to numerical errors. However, recent findings suggest that deep neural networks can operate at the Edge of Stability for an extended, if not the entire, duration of training. Figure \ref{fig:phases} presents a snapshot of a typical training trajectory. \cite{damian2023selfstabilizationimplicitbiasgradient} formalized the notion of progressive sharpening by defining a sharpening parameter, $\alpha=-\nabla L(\theta) \cdot \nabla S(\theta).$\footnote{This can be computed with the Hessian trick.} This is plotted alongside $|\theta - \theta^*|$, the latter estimating the distance to mean oscillation parameters. In examining the trajectory, we delineate distinct phases of learning through the ascent and descent of $|\theta - \theta^*|$. As gradient descent nears the stability boundary of $2/\eta$, parameter oscillations become more pronounced, driving the system toward instability Notably, the peaks in $L(\theta)$ and $S(\theta)$ coincide with peaks in oscillation magnitude. On the other hand, this contrasts with the progressive sharpening factor, $\alpha$, which remains predominantly negative, indicating that minor gradient adjustments often lead to a reduction in $S(\theta)$, contrary to the expected qualitative effects suggested by progressive sharpening and of instability. This evidence suggests that unstable parameter oscillations, rather than progressive sharpening, are critical drivers of instability. 

\begin{figure}[h] 
\includegraphics[width=\linewidth]{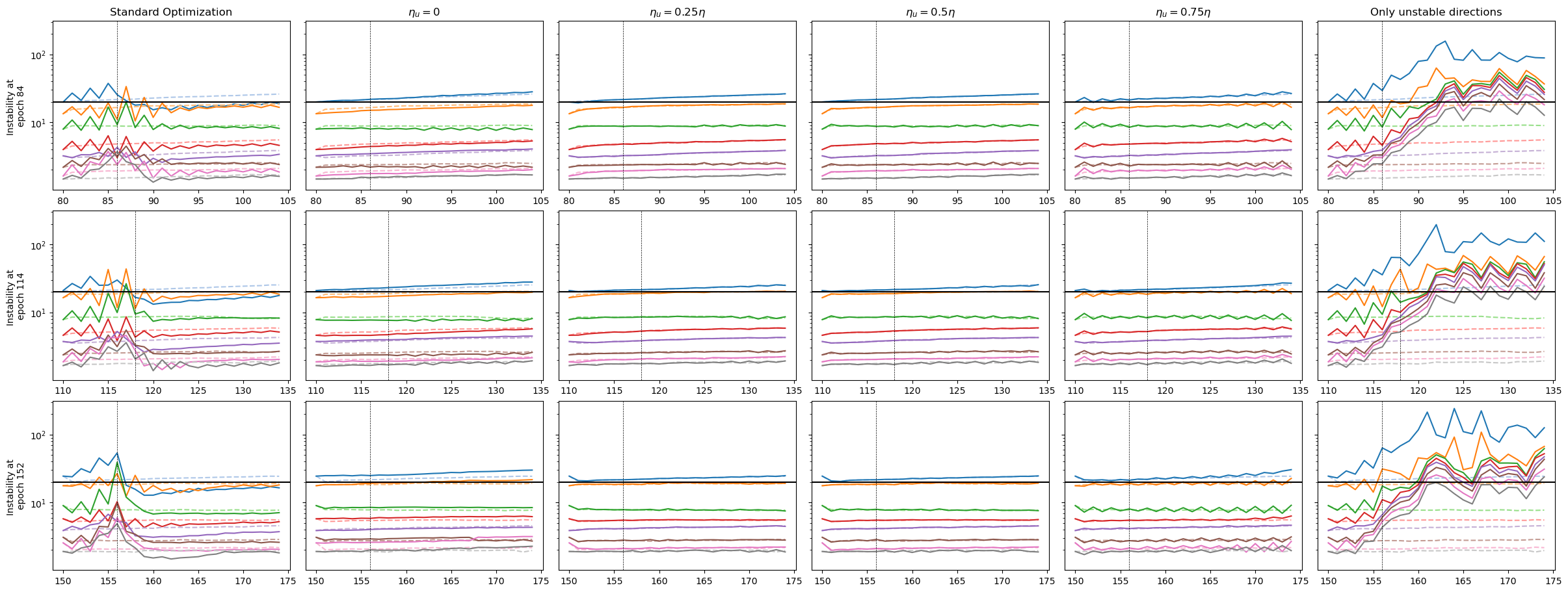}
\caption{\textbf{Limiting step sizes along unstable directions promotes stable training, while restricting updates to only unstable directions leads quickly to instability.} We plot the evolution of $S(\theta)$ and sharpness of the top $8$ eigenvectors on the same instabilities used in Figure \ref{fig:instab_rot}. The stable gradient-flow trajectories are plotted with a dashed line.}
\label{fig:control-pc}
\end{figure}

A more compelling argument arises when we intervene by suppressing parameter updates along unstable directions, setting learning rates to $\eta_u=0$ specifically for these directions. The modified dynamics, shown in Figure \ref{fig:control-pc}, reveal that $S(\theta)$ follows a trajectory similar to that when observed under a globally small learning rate, reflecting a training trajectory completely conducted in the stable regime. This behavior is characterized by slow and gradual increases to $S(\theta)$, which we identify as progressive sharpening, since the latter also exist during stable phases of training. Similarly, reducing the learning rate along unstable directions to achieve \emph{effective stability}, i.e. $\eta_u < 2/S(\theta)$, reproduces this behavior up until the stability threshold is reached, reinforcing the limited role of progressive sharpening toward the formation of instabilities. Conversely, when gradient updates are restricted solely to the directions of the sharpest eigenvectors, unstable oscillations are re-introduced, though the trajectory of $S(\theta)$ significantly changes. 

These findings underscore the fundamental differences between progressive sharpening and unstable oscillations. Progressive sharpening primarily describes the increases in $S(\theta)$, the curvature along the sharpest eigenvector, as a consequence of updates in other directions of the Hessian. In contrast, unstable oscillations are driven specifically by parameter updates along these unstable  directions. Our results demonstrate that unstable parameter oscillations play a critical role in the formation of instabilities, challenging the view that progressive sharpening plays the primary role. This distinction is important to why the generalization performance of deep neural networks is improved when trained with sufficiently large learning rates (which is observed in Section \ref{sec:generalization}), despite the increases in $S(\theta)$ - for all choices of learning rates - due to progressive sharpening. 

\newpage
\section{Complete Evolution of Instability visualized in Figure \ref{fig:movie}} \label{app:movie}

\newpage
\section{Flatness, Rotations, or just Large Learning Rates?} \label{app:gen:exp3-causal}

\begin{figure}[h]
    \centering
    \begin{subfigure}[t]{0.3\textwidth}
        \centering
        \includegraphics[height=3.5cm]{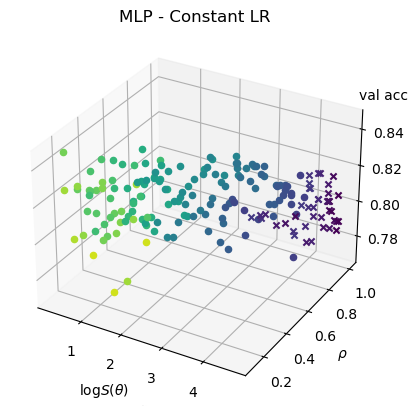}
        \caption{MLP on fMNIST \\ constant $\eta$}
    \end{subfigure}%
    ~ 
    \centering
    \begin{subfigure}[t]{0.3\textwidth}
        \centering
        \includegraphics[height=3.5cm]{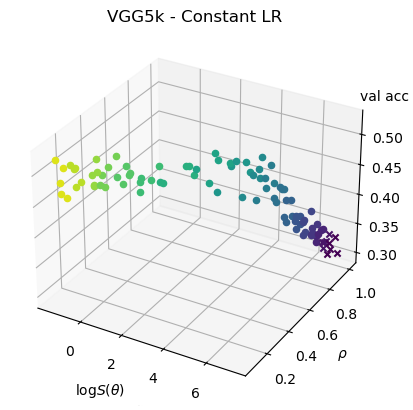}
        \caption{VGG10 on CIFAR10-5k \\ constant $\eta$}
    \end{subfigure}%
    ~ 
    \centering
    \begin{subfigure}[t]{0.3\textwidth}
        \centering
        \includegraphics[height=3.5cm]{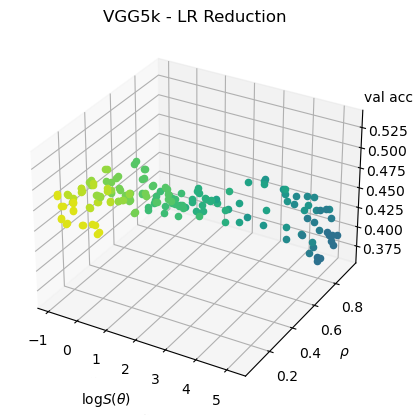}
        \caption{VGG10 on CIFAR10-5k \\ $\eta$ reduction}
    \end{subfigure}%
    ~ 
    \newline
    \centering
    \begin{tabular}{c|c|c|c}
         & $\eta_0$ & $S(\theta)$ & $\rho$ \\
         \hline
         \hline
         MLP on fMNIST; const. $\eta$& 0.595 & -0.593 & \textbf{-0.610} \\
         VGG10 on CIFAR10-5k; const. $\eta$& \textbf{0.9523}& -0.9450& -0.9499 \\
         VGG10 on CIFAR10-5k; $\eta$ reduction & 0.7263 & -0.7763 & \textbf{-0.8200} \\
    \end{tabular}
    
\caption{\textbf{Generalization performance with $S(\theta)$ and $\rho$.} We aggregate models from Figure \ref{fig:goldilocks} and \ref{fig:cifar-flat}, adding an additional dimension $\rho$. Rank correlation with validation error are listed in the table below. }
\label{fig:corr}
\end{figure}

\begin{figure}[t]
    \centering
    \begin{subfigure}[t]{0.6\textwidth}
        \centering
        \includegraphics[width=8cm]{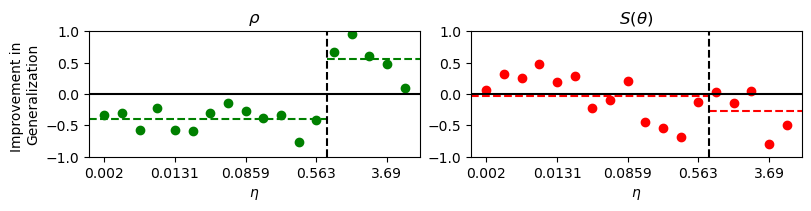}
        \caption{\textbf{Left:} $\rho$, \textbf{Right:} $S(\theta)$. Rank correlation to generalization on sets of $9$ points, a limited sample size. Dashed lines indicate the mean of correlations in each phase. The vertical dotted line separates the above EoS and \emph{extremely-large} phases of $\eta$}
    \end{subfigure}
    ~ 
    ~ 
    \begin{subfigure}[t]{0.3\textwidth}
        \centering
        \begin{tabular}[b]{c|c|c}
             & $\rho$ & $S(\theta)$ \\
             \hline
            \hline
             $\eta$ & \textbf{-0.40} & 0.56\\
             extremely large $\eta$ & -0.03 & \textbf{-0.27}
        \end{tabular}
        \caption{Mean rank correlation \\with generalization}
    
    \end{subfigure}
    \caption{\textbf{Across weight initialization, $\rho$ is a better predictor than $S(\theta)$ of generalization when $\eta$s is controlled. } We group models of the same $\eta$ into subsets of $9$, each differing only by initialization.  }
\label{fig:cifar-kde}
\end{figure}

Our experiments in Sections \ref{sec:gen:exp1-const} and Sections \ref{sec:gen:exp2-reduc} have demonstrated that the generalization performance of models can be improved through the choice of $\eta_0$s. Larger $\eta_0$s bring about models with lower $S(\theta)$ and more rotations. What are the relative importance of these factors, $\eta_0$, $S(\theta)$, and similarity of eigenvectors $\rho$, toward generalization? 


In Figure \ref{fig:corr}, we outline the association with each metric with validation accuracy, where the evidence demonstrates a clear correlation with generalization for each candidate metric. This effect is more pronounced in CIFAR10 than fMNIST, though across these tasks the strengths of these associations are hard to separate under the constant $\eta$ regime, and $\rho$ slightly outperforms alternatives in the $\eta$ reduction regime. With $\eta$ reduction, the models undergo varying degrees of regularization depending on initial learning rates $\eta_0$ and the timing of reduction, with more regularized models tending to prefer stable regimes of training due to progressive flattening. Changes to $S(\theta)$ during stable phases of training are dominated by progressive sharpening, while the orientation of eigenvectors are quickly attracted to fixed points. These differences could explain some of the observed gap. 

In our experiments, we explicitly varied learning rates to indirectly control $S(\theta)$ and $\rho$. This approach is equivalent to assuming the causal precedence of $\eta$, which we can leverage to compare the effects of $S(\theta)$ and $\rho$ on generalization. Assuming no additional confounding factors, the isolated effects of $S(\theta)$ and $\rho$ can be measured by controlling for $\eta$. In Figure \ref{fig:cifar-kde}, we group VGG models trained on CIFAR10-5k into sets of $9$ that share the same learning rates, where randomness is only introduced through weight initialization. The mean correlation of validation accuracy reveals a stronger relationship for $\rho$ than $S(\theta)$, though the overall effect is not conclusive due to the small sample size. In this case, the lack of scale-invariance of $S(\theta)$ limits its ability to compare generalization performance across models when learning rates are controlled for. Our findings highlight a scenario where $S(\theta)$ is not effective, while $\rho$ provides a better predictor of generalization across models with limited evidence.

\end{document}